\documentclass{article}

\usepackage{hyperref}

\usepackage[preprint]{custom}

\usepackage{amsmath,amssymb,amsthm,mathtools}
\usepackage{array,booktabs,multirow,siunitx,makecell}
\usepackage{xurl,graphicx,graphbox,color,pifont,stfloats}
\usepackage{enumitem,caption,subcaption,microtype}
\usepackage[capitalize,noabbrev]{cleveref}
\usepackage[textsize=tiny]{todonotes}
\usepackage[table,dvipsnames]{xcolor}

\customtitlerunning{Kanade: A Simple Disentangled Tokenizer for Spoken Language Modeling}

\newcommand{\NA}{\text{--}}
\begin{document}

\twocolumn[
  \customtitle{Kanade: A Simple Disentangled Tokenizer for Spoken Language Modeling}
  
  \customsetsymbol{equal}{*}
  \begin{customauthorlist}
    \customauthor{Zhijie Huang}{equal,gavo}
    \customauthor{Stephen McIntosh}{equal,gavo}
    \customauthor{Daisuke Saito}{gavo}
    \customauthor{Nobuaki Minematsu}{gavo}
  \end{customauthorlist}
  \customaffiliation{gavo}{The University of Tokyo, Tokyo, Japan}
  \customcorrespondingauthor{Zhijie Huang}{huangzj@gavo.t.u-tokyo.ac.jp}

  \customkeywords{speech tokenization, spoken language modeling, neural audio codec, disentangled representation learning}
  \vskip 0.3in
]
\printAffiliationsAndNotice{\customEqualContribution}

\begin{abstract}
A good language model starts with a good tokenizer. Tokenization is especially important for speech modeling, which must handle continuous signals that mix linguistic and non-linguistic information. A speech tokenizer should extract phonetics and prosody, suppress linguistically irrelevant information like speaker identity, and enable high-quality synthesis. We present Kanade, a single-layer disentangled speech tokenizer that realizes this ideal. Kanade separates out acoustic constants to create a single stream of tokens that captures rich phonetics and prosody. It does so without the need for auxiliary methods that existing disentangled codecs often rely on. Experiments show that Kanade achieves state-of-the-art speaker disentanglement and lexical availability, while maintaining excellent reconstruction quality.
\end{abstract}

\section{Introduction}
Next-token prediction models like GPT can perform various natural language processing tasks without explicit training~\citep{brown2020language}. This has inspired work within spoken language processing to repeat these successes in speech, applying the autoregressive language modeling framework to pure spoken language models (SLMs)~\citep{lakhotia2021on}, text-to-speech (TTS)~\citep{chen2025neural}, and speech-to-speech translation~\citep{lee2022textless}.

In text language models (LMs), the tokenizer splits text into subword units. In autoregressive speech models, the speech tokenizer plays a similar, but more demanding role~\citep{mousavi2025discrete}. 
As with text, good speech representations should surface the basic units of language~\citep{borsos2023audiolm, guo2025recent}, from which we can recover higher-level features like morphology, syntax, and pragmatics. Importantly, these should include both phonetic and prosodic (intonation, stress, and rhythm) information~\citep{kharitonov2022text}.
Unlike text, which is already a semantically dense discrete representation of human language, recordings of speech are continuous waveforms that also include linguistically irrelevant information such as background noise and speaker identity. This makes extracting meaningful discrete representations a difficult task.

\begin{figure*}[t]
\centering
\includegraphics[width=0.8\linewidth]{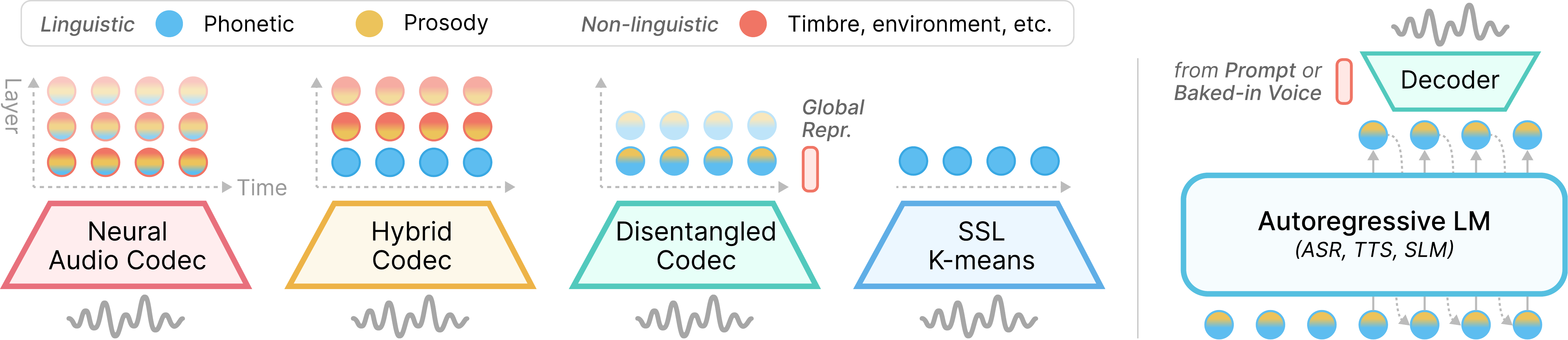}
\caption{\textbf{Left:} Comparison of information distribution in major classes of speech tokenizers. Kanade belongs to the class of single-layer disentangled codec. \textbf{Right:} Usage of Kanade tokens for various speech LM tasks. Color gradients represent mixed content. Adapted from SpeechTokenizer~\citep{zhang2024speechtokenizer}.}
\label{fig:concept}
\end{figure*}

One of the earliest methods for speech tokenization uses k-means clustered features from self-supervised learning (SSL) models~\citep{lakhotia2021on}. These SSL tokens capture phonetic structure well but discard necessary prosodic information~\citep{kharitonov2022text}. In contrast, neural audio codecs (NACs) generally retain too much acoustic variance in their heavy multi-layer representations, requiring downstream models to learn complex distributions (see Figure~\ref{fig:concept}, left). This was observed by \citet{borsos2023audiolm}, who proposed to combine the strengths of these tokens by using SSL tokens for language modeling and then converting those to NAC tokens in a separate step. This design allows the main language model to focus on phonetic content, but then uses a different model to fill in acoustic details. This produces high-quality coherent speech, but requires a complex architecture.

An ideal speech tokenizer should organize information in a way that is conducive to downstream modeling~\citep{dunbar2022self}. It should:
{
\parskip=0pt
\begin{description}[leftmargin=0.5em, nolistsep, topsep=0pt, partopsep=0pt]
    \item[Include phonetic and prosodic information] The meaning of speech is mostly determined by its phonetic content, but prosody is also essential to human communication~\citep{cutler1997prosody, dahan2015prosody}.
    The main advantage SLMs have over text LM cascades is that they can use prosodic features for better comprehension and to generate expressive speech~\citep{kharitonov2022text}.

    \item[Suppress non-linguistic information] Downstream models can learn more efficiently if we provide them with representations that contain only relevant information~\citep{tishby2015deep, vandenoord2017neural}. This idea is similar to how image encoders are often optimized to produce representations that encode the identity of the pictured object rather than orientation, lighting, or camera characteristics.

    \item[Enable high-quality synthesis]
    Synthesizing high-quality speech requires both the linguistic information contained in the tokens and non-linguistic information. The tokenizer should provide an easy way to inject the missing information (e.g., speaker identity) when decoding tokens back to speech~\citep{borsos2023audiolm, guo2025recent}.
\end{description}
}

These requirements naturally motivate the design of tokenizers that separate linguistic and non-linguistic information. Disentangled codecs (see Figure~\ref{fig:concept}, left) often realize this by separating speech into time-varying content and acoustic invariants. Since many linguistically irrelevant features like speaker identity and microphone characteristics are constant, this allows the content stream to focus on linguistics, while relegating information necessary for reconstruction to a separate representation (see Figure~\ref{fig:concept}, right). \citet{martin2024enhancing} have shown that downstream language models that use only the content stream perform better since they learn only the content distribution, rather than a more complicated joint distribution mixed with non-linguistic detail.

However, disentangled codecs often resort to auxiliary methods to enforce separation, such as gradient reversal~\citep{ju2024naturalspeech}, invariance learning~\citep{martin2024enhancing, ren2024fewer} or supervision~\citep{khurana2025factorized}. Even with these methods, our empirical analysis reveals that current implementations often yield poor disentanglement.

In this work, we present Kanade, a disentangled single-layer speech tokenizer. Kanade uses only a narrow information bottleneck to achieve clean unsupervised disentanglement, eschewing the above methods. Kanade produces a single stream of 12.5/25 Hz discrete tokens that are rich in phonetics and prosody. This single stream can be modeled without complex downstream architectures, and the low token rate is suitable for autoregressive models. Kanade requires only 600 hours of data to train and 120M unfrozen parameters, but achieves state-of-the-art (SOTA) metrics among codecs on (1) \textbf{speaker disentanglement}, as measured by voice conversion and discrimination tasks, and (2) \textbf{lexical availability}, as measured by downstream automatic speech recognition (ASR) and TTS. In our pure SLM experiments, Kanade achieves performance competitive with SSL tokens. At the same time, it maintains excellent reconstruction quality comparable with multi-layer codecs and enables superior prosodic naturalness in TTS generation.

Thanks to its effective disentanglement, Kanade combines the linguistic availability of SSL tokens and the generation quality of NACs in a single-layer token stream. It achieves this using a simple, well-motivated architecture:
{
\parskip=0pt
\begin{itemize}[leftmargin=1em, nolistsep, topsep=0pt, partopsep=0pt]
    \item \textbf{Uses only SSL features as input}. SSL features have been shown to be sufficient for faithful reconstruction~\citep{zhang2025vevo} and possess a structured latent space where content and speaker information is easily separable~\citep{kamper2025linearvc}. This gives Kanade a head-start on disentanglement and improves data efficiency.
    \item \textbf{Uses both SSL and audio reconstruction loss}. The SSL feature space is sensitive to phonetic contrasts, whereas the acoustic space is sensitive to prosody. Using reconstruction losses on both helps the tokenizer encode maximal linguistic information~\citep{ye2025codec}.
    \item \textbf{Uses codebook-free quantization}, allowing it to effectively quantize the content into a single layer~\citep{mentzer2024finite}.
\end{itemize}
}

Our contributions:
{
\parskip=0pt
\begin{itemize}[leftmargin=1em, nolistsep, topsep=0pt, partopsep=0pt]
    \item We provide a simple, open-source recipe\footnote{\url{https://github.com/frothywater/kanade-tokenizer}} for a single-layer speech tokenizer that achieves best-in-class disentanglement and lexical availability without auxiliary methods, while maintaining excellent prosody preservation and reconstruction quality.
    \item We assemble a comprehensive suite of metrics to benchmark a variety of open-source speech tokenizers.
    \item We demonstrate that a single-layer disentangled codec can provide the desirable properties of SSL tokens and NAC tokens without requiring complicated downstream architectures.
\end{itemize}
}

\section{Related Work}

\textbf{SSL tokens} are derived from SSL models such as wav2vec 2.0~\citep{baevski2020wav2vec}, HuBERT~\citep{hsu2021hubert}, and WavLM~\citep{chen2022wavlm}. These models typically capture accessible phonetic~\citep{pasad2023comparative} and prosodic~\citep{chiu2025large} information, as well as easily separable speaker attributes~\citep{kamper2025linearvc}. Most pure SLMs utilize these representations by discretizing them via k-means~\citep{lakhotia2021on}. Unfortunately, k-means tokens largely discard prosodic information, making them unsuitable for prosody modeling and resynthesis~\citep{kharitonov2022text}. RepCodec~\citep{huang2024repcodec} shows that quantizing SSL features with VQ-VAE~\citep{vandenoord2017neural} improves prosody preservation. We extend this by including audio reconstruction loss to enhance the performance.

\textbf{NACs} such as EnCodec~\citep{defossez2023high} are designed to compress audio and trained with reconstructive losses, leading to high information preservation suitable for resynthesis~\citep{ji2024wavtokenizer, parker2024scaling}. However, NACs based on Residual Vector Quantization (RVQ) typically produce multiple tokens at each timestep, which results in high token rates and often obscures the underlying linguistic structure~\citep{borsos2023audiolm, mousavi2024dasb}. 

\textbf{Hybrid codecs} such as SpeechTokenizer~\citep{zhang2024speechtokenizer} seek to bridge this gap by enhancing phonetic information of NACs, typically by distilling SSL features into their first RVQ layer~\citep{defossez2024moshi, zheng2024freecodec} or using them directly as inputs~\citep{ye2025codec, li25dualcodec}. However, these methods often rely on the multi-layer token structures, which complicate downstream usage and potentially reduce efficiency~\citep{borsos2023audiolm, guo2025recent}. Furthermore, although hybrid codecs exhibit a degree of separation between linguistic and non-linguistic information~\citep{zhang2024speechtokenizer}, our experiments show that this separation is often incomplete: due to the lack of explicit disentanglement, linguistic content tends to leak into higher token layers.

\textbf{Disentangled codecs} such as FACodec~\citep{ju2024naturalspeech} use explicit architectures to enforce speech disentanglement, often combined with additional losses and techniques. Common auxiliary methods include adversarial learning (specifically gradient reversal)~\citep{ju2024naturalspeech}, contrastive learning~\citep{martin2024enhancing, vecino2025universal}, invariance learning~\citep{ren2024fewer, li2024single, guo2024socodec, guo2024lscodec, zheng2024freecodec}, and supervision~\citep{khurana2025factorized}. These can complicate the training pipeline and reduce scalability. Moreover, our experiments suggest they do not necessarily lead to effective disentanglement, as non-linguistic features often leak into content tokens. 

In contrast, Kanade achieves unsupervised disentanglement by only applying information bottleneck in a two-branch architecture and eliminates the need for multi-layer token structures. To our knowledge, only BiCodec~\citep{wang2025spark} has attempted a similar strategy. However, BiCodec uses a more complicated global branch, and our voice conversion experiments demonstrate that its disentanglement is not ideal. See Appendix~\ref{appendix:model-comparison} for a detailed comparison to other works.

\section{Method}
\begin{figure}[t]
\centering
\includegraphics[width=\linewidth]{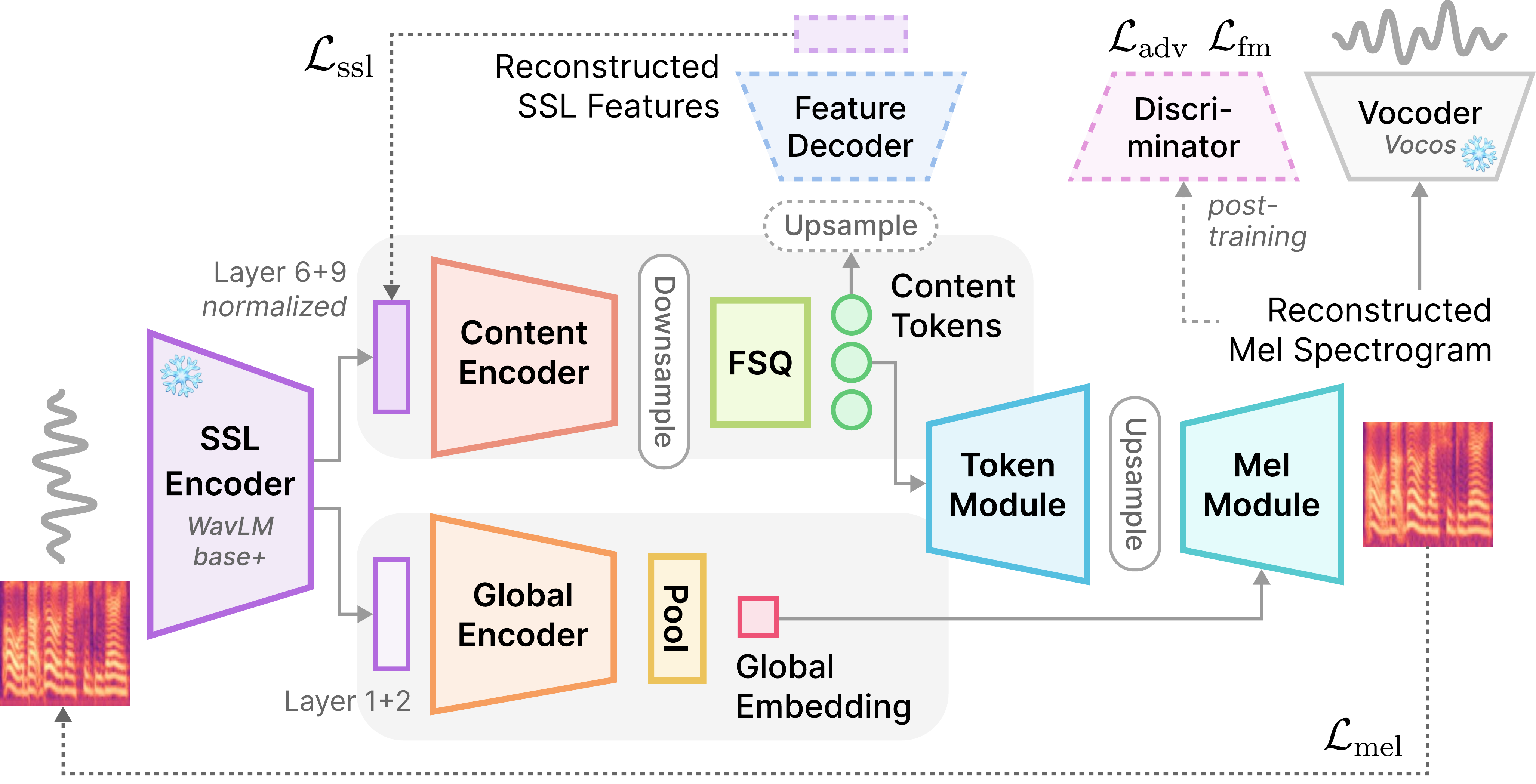}
\caption{\textbf{Model architecture of Kanade}}
\label{fig:model}
\end{figure}

The architecture of Kanade is illustrated in Figure~\ref{fig:model}. First, we use an \hyperref[section:ssl-encoder]{SSL encoder} to extract SSL features from various layers. Features from deep layers, associated with linguistic content~\citep{pasad2023comparative, chiu2025large}, are fed into a \textit{content} branch (top gray box, Section~\ref{section:content_branch}) which further encodes the speech and then quantizes it into tokens (green circles). Features from shallow layers, associated with speaker characteristics~\citep{chen2022why}, are fed into a \textit{global} branch (bottom gray box, Section ~\ref{section:global-branch}) which produces a single continuous embedding (red square). The decoder (right side of Figure~\ref{fig:model}, Section~\ref{section:decoder}) reconstructs the waveform from the content tokens and global embedding.  We train using SSL feature and mel spectrogram reconstructions losses (Section~\ref{section:main-losses}), then perform GAN post-training to improve audio quality (Section ~\ref{section:gan-losses}).
To summarize our approach:
{
\parskip=0pt
\begin{description}[leftmargin=1em, nolistsep, topsep=0pt, partopsep=0pt]
    \item[SSL reconstruction loss] on content-rich SSL features emphasizes phonetic information.
    \item[Mel reconstruction loss] is sensitive to prosodic features, so encourages the content branch to include them.
    \item[A global branch] provides a path for non-linguistic information to flow through. SSL feature reconstruction loss is relatively insensitive to this information, so the bitrate-constrained content encoder is encouraged to drop it.
\end{description}
}

\subsection{Architecture}
\textbf{SSL encoder}\phantomsection\label{section:ssl-encoder}\; SSL features already contain the information that we would like to extract from speech, including not only phonetics and prosody~\citep{zhang2025vevo}, but also easily separable reconstruction-related information in a structured latent space~\citep{kamper2025linearvc}. Therefore, it is more effective to reconfigure these representations than start with the raw audio or mel spectrogram. We use WavLM Base+~\citep{chen2022wavlm} as our SSL encoder and freeze it during training (we also try another encoder in Appendix~\ref{appendix:ablation-arch} and observe similar results). See Appendices~\ref{appendix:ablation-content} and \ref{appendix:ablation-global} for layer selection ablations.

\subsubsection{Content branch}\label{section:content_branch}
\textbf{Content encoder}\; We average the content layers' representations (Layer 6 and 9) and normalize each dimension to zero mean and unit variance. We pass these features through a transformer encoder, selected for its strong modeling ability (see Appendix~\ref{appendix:ablation-arch} for an ablation over CNN). Following \citet{parker2024scaling}, we use local window attention in all our transformers for locality bias and computational efficiency. The encoder outputs are temporally downsampled via a strided convolution.

\textbf{Vector quantization}\; We use a VQ-VAE~\citep{vandenoord2017neural} architecture for extracting discrete tokens given its success in prior work~\citep{defossez2023high, huang2024repcodec}. Unfortunately, the vector quantization method used by \citet{vandenoord2017neural} is sensitive to initialization, prone to codebook collapse, and can have difficulty keeping up with constantly moving encoder outputs~\citep{lancucki2020robust, zhu2024addressing}. Though previous work uses RVQ~\citep{gray1984vector} to alleviate these problems, we wanted to produce one token per timestep so opted to use Finite Scalar Quantization (FSQ)~\citep{mentzer2024finite} to quantize encoder outputs. FSQ is a simple codebook-free method and avoids many of the problems caused by a dynamic codebook (see the ablation in Table~\ref{table:ablation-main}).
To obtain tokens, representations from the content encoder are projected to the FSQ dimension, quantized, and represented by their indices in the implied codebook.

\subsubsection{Global branch}\label{section:global-branch}
Linguistic information can only be conveyed by time-varying features. The goal of the global branch is to capture time-invariant information, providing a path for non-linguistic information to flow through, so that the bitrate-constrained content branch can focus on linguistic content (see the ablation in Table~\ref{table:ablation-main}). To this end, we produce a single global embedding for the entire utterance.

The global branch architecture is inspired by NeXt-TDNN~\citep{heo2024nexttdnn}, an updated version of ECAPA-TDNN~\citep{desplanques2020ecapa} architecture with modified ConvNeXt~\citep{liu2022convnet} blocks. We did not use a transformer for the global branch because there is no long-term dependency that we would like to capture. The global embedding is not discretized because we don't expect it to be used in autoregressive modeling and our focus is learning high-quality content tokens. \citet{li2024single} showed that discretizing it may be detrimental. Furthermore, we show in Appendix~\ref{appendix:global-embedding-pca} that the continuous representation can be freely manipulated to condition the decoder.

The shallow SSL representations for the global branch (Layer 1 and 2) are averaged, but not normalized. They are then passed to the \textbf{Global Encoder}, which is a stack of standard ConvNeXt blocks.
To obtain a single embedding for the entire sequence, we use an \textbf{attentive statistics pooling}~\citep{okabe2018attentive}, following ECAPA-TDNN. An ablation using average pooling instead shows this slightly improves reconstruction quality (see Appendix~\ref{appendix:ablation-global}).

\subsubsection{Decoder}\label{section:decoder}
The first step of decoding is to convert the content tokens back into vectors in the FSQ implied codebook.
These are passed through two transformer-based decoder modules: the \textbf{Token Module} and \textbf{Mel Module}, inspired by TTS systems~\citep{ren2021fastspeech}.
Since our tokens are produced at a constant rate, we use transposed strided convolution to upsample features before feeding them to the mel module instead of a duration predictor.

The mel module's role is to produce a final mel spectrogram. It is conditioned by the global embedding using adaLN-Zero~\citep{peebles2023scalable}. All timesteps receive the same conditioning. We choose adaptive layer normalization based on its success in AdaSpeech~\citep{wu2022adaspeech} and use the zero variant because it has better training characteristics. 
A convolutional post-net is applied at the end to refine the generated spectrogram.
We target a mel spectrogram rather than a waveform mainly to simplify training. The focus of our work is token quality and we found it sufficient to use Vocos~\citep{siuzdak2024vocos} as a final step to convert the mel spectrogram to a waveform.

\subsection{Training objectives}
\subsubsection{Main training phase}\label{section:main-losses}
\textbf{Feature reconstruction}\; Since the SSL representations that the content branch uses surface useful phonetic information~\citep{pasad2023comparative}, we use a feature reconstruction loss to preserve that information in our tokens. Ablation shows this is very important, as seen in Table~\ref{table:ablation-main}. To compute this, we convert the tokens back into vectors and upsample with a transposed strided convolution to the SSL frame rate. We then pass these to the transformer-based \textbf{Feature Decoder} to reconstruct the input. We compare the results with the input to the content encoder and compute the L2 loss $\mathcal L_\text{ssl}$, as was done by RepCodec~\citep{huang2024repcodec}. The feature decoder is used in training only.

\textbf{Mel reconstruction}\; We compute L1 loss from the reconstructed mel spectrogram to obtain $\mathcal L_\text{mel}$, following convention~\citep{kim2021conditional}.

We combine these two losses to obtain $\mathcal L = \mathcal L_\text{mel} + \alpha \mathcal L_\text{ssl}$ in the main training phase. We did not observe high sensitivity of the weight $\alpha$ in our preliminary experiments so fixed the value to 1. We also tried splitting the losses into two stages, with only SSL loss at first, then switching to mel reconstruction loss; however, we found this caused the encoder to ignore some prosodic features (see the ablation in Table~\ref{table:ablation-main}). This is similar to how k-means (which is also computed using L2 distances in SSL feature space) loses prosodic information before phonetic information~\citep{kharitonov2022text, onda2025benchmarking}, so we suspect that distances in phonetically rich SSL layers are comparatively less sensitive to prosodic features.
Therefore, including mel reconstruction loss encourages the encoder to extract richer prosodic information.

\subsubsection{GAN post-training}\label{section:gan-losses}
With only the main training phase, the model produces intelligible speech (see the ablation in Appendix~\ref{appendix:ablation-arch}), but the spectrogram is blurry, degrading audio quality. \citet{wu2023audiodec} show that introducing GAN~\citep{goodfellow2014generative} post-training on the decoder can restore finer details.
To avoid passing gradients through the vocoder, we compare the mel spectrograms rather than the waveforms, using a multi-band discriminator design similar to DAC~\citep{kumar2023high}.
We use adversarial loss $\mathcal L_\text{adv}$ and feature matching loss $\mathcal L_\text{fm}$ as described in Vocos~\citep{siuzdak2024vocos}. During post-training, only the global branch and the decoder are updated.
The post-training objective is $\mathcal L_\text{post} = \mathcal L_\text{mel} + \beta \mathcal L_\text{adv} + \gamma \mathcal L_\text{fm}$.

\section{Experiments}

\subsection{Training setup}
\label{section:data}

The resulting models have $\sim$120M trainable parameters and $\sim$210M total parameters (containing WavLM Base+ and Vocos). Training of two phases takes approximately 32 hours on one NVIDIA 5090 GPU.
Details on our model and training configurations can be found in Appendix~\ref{appendix:model} and \ref{appendix:efficiency}. We train our models using all training sets of LibriTTS~\citep{zen2019libritts}, a multi-speaker English corpus containing 586 hours of audiobook speech sampled at 24kHz. LibriTTS is derived from the same materials as the LibriSpeech~\citep{panayotov2015librispeech} corpus.

\subsection{Baselines} \label{baselines}
We compare Kanade with a variety of SOTA speech codecs, including single-layer codecs, hybrid codecs, and disentangled codecs. See Appendix~\ref{appendix:baselines} for more details. SpeechTokenizer~\citep{zhang2024speechtokenizer} is abbreviated as ST.

We also train several reference models that change the way content is encoded. We train k-means reference models (KM) that use the same SSL representations used by the content encoder (see Section~\ref{section:content_branch}---normalizing before clustering is consistent with prior work~\citep{borsos2023audiolm}). These features are downsampled with average pooling and clustered using k-means, which is trained on the LibriTTS train subsets. Note that k-means models use global embeddings as our main models do. A separate continuous reference model (Cont.) is trained by replacing both encoding branches with full-resolution (50Hz) continuous SSL features. Since we remove the global branch, these are an average of all four layers used in our main models.

\begin{table}[t]
\caption{\textbf{Speech reconstruction results.} The top group includes \colorbox{gray!10}{reference metrics}. Only models that are best in some metric are included. The bold numbers are the best in their group. For all results, see Table~\ref{table:reconstruction-all} in the appendix. For MUSHRA confidence intervals, see Table~\ref{table:reconstruction-mushra}. WER and CER are in percentage (\%).}
\label{table:reconstruction}
\scriptsize
\centering
\setlength{\tabcolsep}{1pt}
\begin{tabular}{
    m{6.5em} 
    >{\centering\arraybackslash}m{2.5em} 
    S[table-format=2.1,detect-weight]
    S[table-format=2.1,detect-weight]
    S[table-format=2.2,detect-weight]
    S[table-format=1.2,detect-weight]
    S[table-format=1.2,detect-weight]
    S[table-format=1.2,detect-weight]
}
\toprule
\multirow{2}{*}{\textbf{Model}} & \multirow{2}{*}{\makecell{\textbf{Token} \\ \textbf{Rate}}} & \multicolumn{2}{c}{\textbf{Intelligibility}} & \multicolumn{2}{c}{\textbf{Quality}} & \multicolumn{1}{c}{\textbf{Speaker}} & \multicolumn{1}{c}{\textbf{Prosody}} \\
\cmidrule(lr){3-4} \cmidrule(lr){5-6} \cmidrule(lr){7-7} \cmidrule(lr){8-8}
& & {WER$\downarrow$} & {CER$\downarrow$} & {MUSHRA$\uparrow$} & {UTMOS$\uparrow$} & {SIM$\uparrow$} & {F0Corr$\uparrow$} \\
\midrule
\rowcolor{gray!10} Ground Truth & {--} & 1.9 & 0.6 & 78.0 & 4.07 & {--} & {--} \\
\rowcolor{gray!10} Cont. 50Hz & {--} & 2.0 & 0.6 & 72.1 & 3.90 & 0.99 & 0.94 \\
\rowcolor{gray!10} KM 12.5Hz & 12.5 & 3.0 & 1.1 & 72.1 & 4.04 & 0.96 & 0.66 \\
\rowcolor{gray!10} KM 25Hz & 25 & 2.7 & 1.0 & 72.4 & 4.07 & 0.96 & 0.67 \\
\cmidrule(lr){1-8}
\noalign{\vskip -1pt}
\multicolumn{8}{c}{Multi-layer} \\
\noalign{\vskip -1pt}
\cmidrule(lr){1-8}
FACodec & 480 & \bfseries 2.1 & \bfseries 0.7 & 81.4 & 4.11 & 0.98 & 0.94 \\
PAST & 400 & \bfseries 2.1 & \bfseries 0.7 & \bfseries 82.4 & \bfseries 4.18 & \bfseries 0.99 & 0.92 \\
ST & 400 & \bfseries 2.1 & \bfseries 0.7 & 76.0 & 3.90 & 0.98 & 0.92 \\
DualCodec & 100 & \bfseries 2.1 & \bfseries 0.7 & 75.6 & 4.12 & 0.98 & \bfseries 0.95 \\
\cmidrule(lr){1-8}
\multicolumn{8}{c}{Single-layer} \\
\noalign{\vskip -1pt}
\cmidrule(lr){1-8}
X-Codec 2 & 50 & 2.5 & 0.9 & 77.0 & 4.13 & \bfseries 0.98 & 0.90 \\
BiCodec & 50 & 2.5 & 0.9 & 75.0 & 4.18 & \bfseries 0.98 & \bfseries 0.91 \\
WavTokenizer & 40 & 9.4 & 4.7 & 72.1 & 3.57 & 0.92 & \bfseries 0.91 \\
StableCodec & 25 & 5.7 & 2.6 & \bfseries 79.3 & \bfseries 4.31 & 0.93 & \bfseries 0.91 \\
\cmidrule(lr){1-8}
Kanade 12.5Hz & 12.5 & 3.3 & 1.3 & 74.6 & 4.17 & 0.97 & 0.85 \\
Kanade 25Hz & 25 & \bfseries 2.4 & \bfseries 0.8 & 75.0 & 4.16 & 0.97 & 0.88 \\
\bottomrule
\end{tabular}
\end{table}

\subsection{Evaluation}
\label{section:metrics}
We evaluate generated speech according to:
(1) \textbf{intelligibility}: word/character error rate (WER/CER) using Parakeet \footnote{\url{https://huggingface.co/nvidia/parakeet-tdt-0.6b-v3}};
(2) \textbf{quality}: MUSHRA\footnote{
More details about the listening test are in Appendix~\ref{appendix:subjective}.
} and UTMOS~\citep{saeki2022utmos};
(3) \textbf{speaker identity}: speaker embedding cosine similarity (SIM) using WavLM Base+ for Speaker Verification\footnote{\url{https://huggingface.co/microsoft/wavlm-base-plus-sv}} (WavLM-SV); and
(4) \textbf{prosody}: log F0 Pearson correlation (F0Corr), extracted by SWIPE~\citep{camacho2008sawtooth}.
Evaluation code is largely adapted from VERSA~\citep{shi2025versa}.
We also evaluate our models and baselines using various downstream tasks. The relevant metrics will be introduced along with their results.

\subsection{Results}

\begin{figure}[t]
\centering
\includegraphics[width=\linewidth,trim={0.4cm 0.4cm 0.3cm 0.3cm},clip]{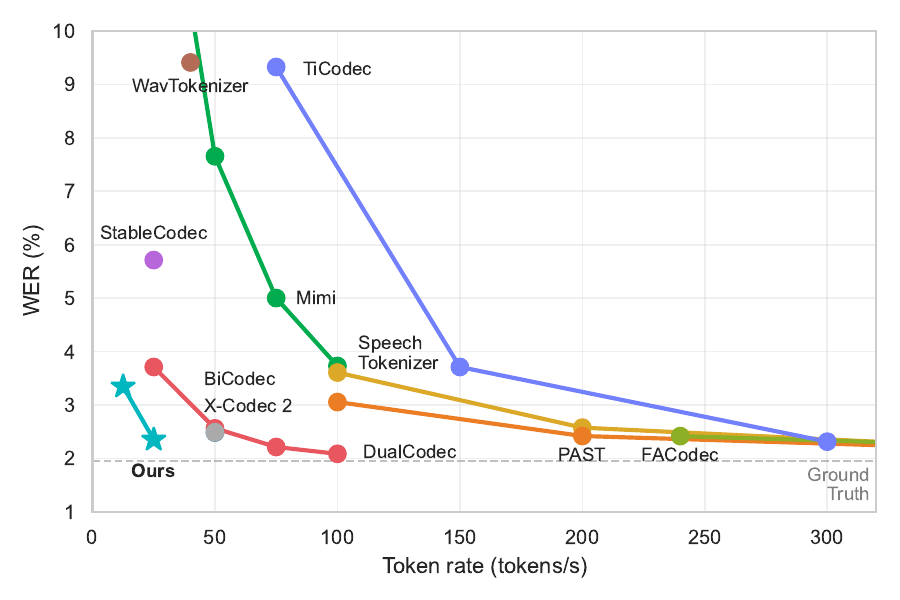}
\caption{\textbf{Reconstruction lexical accuracy (WER) vs. token rate}}
\label{fig:wer-rate}
\end{figure}

\subsubsection{Reconstruction}
\label{section:reconstruction}

We evaluate speech reconstruction on LibriSpeech \texttt{test-clean}. The results are shown in Table~\ref{table:reconstruction}.
Kanade maintains high speech quality, achieves the best WER among single-layer codecs and even approaches the heaviest RVQ models (see Figure~\ref{fig:wer-rate}), demonstrating its token efficiency for linguistic information. The k-means reference models have significantly degraded prosody preservation (0.67 KM 25Hz vs. 0.88 Kanade 25Hz on F0Corr), even when conditioned by the global embedding. This indicates that our content tokens capture prosodic information better than k-means tokens, which is further confirmed by probing (Appendix~\ref{appendix:prosody}).
Evaluation on noisy, spontaneous, emotional, accented and unseen language data reveals Kanade's competitive robustness (Appendix~\ref{appendix:ood}). Also, Kanade has good length generalization and high inference speed; we prototype chunked streaming to demonstrate its practical usage (Appendices~\ref{appendix:length-gen}, \ref{appendix:efficiency}, and \ref{appendix:chunked-encoding-decoding}).

\subsubsection{Disentanglement}

\newcommand{\purplenarrow}{\mathrel{\scalebox{0.9}{\textcolor{Blue}{$\nrightarrow$}}}}
\newcommand{\redneq}{\mathrel{\scalebox{0.8}{\textcolor{Maroon}{$\neq$}}}}

\textbf{Voice Conversion (VC)}\; To measure disentanglement in hybrid and disentangled speech tokenizers, we combine content tokens (usually RVQ layer 1 in hybrid codecs) extracted from \textit{source} utterances with remaining tokens and embeddings extracted from \textit{reference} (or \textit{target}) utterances and then resynthesize. This is done for 1,000 gender-balanced (source, reference) pairs from VCTK~\citep{yamagishi2019cstrvctk}. We randomly select 20 source speakers, 10 target speakers, and 5 source sentences.

If the phonetics and prosody match the \textit{source}, and the timbre matches the \textit{reference}, this indicates good disentanglement.
Linguistic content is measured using WER and prosodic correlation (F0Corr) with respect to the \textit{source}. 
To measure speaker similarity, we use WavLM-SV to discriminate real and converted speech from a \textit{target} speaker following \citet{das2020predictions}. A higher equal error rate (EER) indicates higher difficulty for the system to discriminate real and fake speech, suggesting better timbre transfer.
We additionally conduct MUSHRA-like listening tests to subjectively evaluate speaker similarity (see Appendix~\ref{appendix:subjective}).
Specialized VC models were included as baselines: LinearVC~\citep{kamper2025linearvc}, FreeVC~\citep{li2023freevc}, and CosyVoice 2~\citep{du2024cosyvoice2scalablestreaming}.

Results are shown in Table~\ref{table:vc}. We observe two failure patterns in baselines: (1) \textbf{Poor timbre transfer} ($\purplenarrow$) in disentangled codecs (EER$<$20\%), suggesting non-linguistic information leaking into content tokens; in subjective listening, we often notice gender mixing. (2) \textbf{Content degradation} ($\redneq$) in hybrid codecs (WER$>$20\%, F0Corr$<$0.60), indicating linguistic content leaking into higher layers; in the extreme case, Mimi, we notice that the original content is nearly lost and the reference speech is reconstructed. Samples of all the models can be found at our demo site\footnote{\url{https://frothywater.github.io/kanade-tokenizer/}}.

Among the tested models, Kanade is the only speech codec that both preserves content and achieves effective timbre transfer. Moreover, our performance matches or even surpasses specialized VC models, demonstrating that our simple architecture achieves excellent disentanglement.

\begin{table}[t]
\centering
\caption{\textbf{Voice conversion results.} Bold numbers are the best among codecs. $\purplenarrow$ and $\redneq$ denote models exhibiting poor timbre transfer and serious content degradation, respectively. For the full results, see Table~\ref{table:vc-full} and \ref{table:vc-mushra}.}
\label{table:vc}
\scriptsize
\begin{tabular}{
    m{7em} 
    S[table-format=3.1,round-mode=places,round-precision=1,detect-weight]
    S[table-format=2.1,round-mode=places,round-precision=1,detect-weight]
    S[table-format=1.2,round-mode=places,round-precision=2,detect-weight]
    S[table-format=2.1,round-mode=places,round-precision=1,detect-weight]
    S[table-format=2.1,round-mode=places,round-precision=1,detect-weight]
    S[table-format=1.2,round-mode=places,round-precision=2,detect-weight]
}
\toprule
\multirow{2}{*}{\textbf{Model}} & 
\multicolumn{2}{c}{\textbf{Lexical Content}} & 
\multicolumn{1}{c}{\textbf{Quality}} & 
\multicolumn{2}{c}{\textbf{Speaker Timbre}} & 
\multicolumn{1}{c}{\textbf{Prosody}} \\
\cmidrule(lr){2-3} \cmidrule(lr){4-4} \cmidrule(lr){5-6} \cmidrule(lr){7-7}
& {WER$\downarrow$} & {CER$\downarrow$} & {UTMOS$\uparrow$} & {EER$\uparrow$} & {Similarity$\uparrow$} & {F0Corr$\uparrow$} \\
\midrule
\rowcolor{gray!10} Ground Truth & 0.0 & 0.0 & 4.084 & \NA & \NA & \NA \\
\rowcolor{gray!10} LinearVC & 0.6 & 0.2 & 3.941 & 29.7 & 73.4 & 0.621 \\
\rowcolor{gray!10} FreeVC & 0.6 & 0.3 & 3.993 & 29.0 & 74.5 & 0.673 \\
\rowcolor{gray!10} CosyVoice 2 & 1.1 & 0.5 & 4.108 & 31.0 & 76.0 & 0.638 \\
\midrule
TiCodec \hfill $\purplenarrow$ & \bfseries 0.5 & \bfseries 0.2 & 3.318 & 5.4 & 68.0 & 0.772 \\
FACodec \hfill $\purplenarrow$ & 0.7 & 0.4 & 3.765 & 15.2 & 62.6 & \bfseries 0.789 \\
BiCodec \hfill $\purplenarrow$ & 1.2 & 0.6 & 3.844 & 18.5 & 71.4 & 0.608 \\
DualCodec \hfill $\purplenarrow\redneq$ & 21.5 & 12.9 & 2.505 & 6.8 & 52.0 & 0.542 \\
PAST \hfill $\purplenarrow\redneq$ & 22.9 & 15.1 & 1.841 & 8.2 & 23.3 & 0.200 \\
ST \hfill $\purplenarrow\redneq$ & 74.7 & 61.7 & 1.544 & 10.6 & 35.0 & 0.195 \\
Mimi \hfill $\redneq$ & 120.3 & 86.8 & 3.086 & \bfseries 38.5 & \bfseries 81.7 & 0.213 \\
\midrule
Kanade 12.5Hz & 1.6 & 0.7 & \bfseries 4.167 & 32.0 & 77.6 & 0.640 \\
Kanade 25Hz & 0.7 & 0.3 & 4.156 & 30.7 & 77.1 & 0.707 \\
\bottomrule
\end{tabular}
\end{table}
\begin{table}
\caption{\textbf{Downstream speaker discrimination results (\%) on different parts of disentangled codec representations}}
\label{table:disentangle-downstream}
\scriptsize
\centering
\begin{tabular}{
    m{6.5em}
    S[table-format=2.1,round-mode=places,round-precision=1,detect-weight]
    S[table-format=2.1,round-mode=places,round-precision=1,detect-weight]
    S[table-format=2.1,round-mode=places,round-precision=1,detect-weight]
    S[table-format=2.1,round-mode=places,round-precision=1,detect-weight]
    S[table-format=2.1,round-mode=places,round-precision=1,detect-weight]
    S[table-format=2.1,round-mode=places,round-precision=1,detect-weight]
}
\toprule
\multirow{2}{*}{\textbf{Model}} & 
\multicolumn{3}{c}{\textbf{SID Acc}$\uparrow$} & 
\multicolumn{3}{c}{\textbf{ASV EER}$\downarrow$} \\
\cmidrule(lr){2-4} \cmidrule(lr){5-7}
& {Content} & {Global} & {Both} & {Content} & {Global} & {Both} \\
\midrule
FACodec & 76.8 & 0.3 & 64.7 & 8.9 & 37.0 & 11.8 \\
TiCodec & 56.2 & 4.3 & 23.9 & 13.4 & 43.0 & 20.4 \\
BiCodec & 0.5 & 27.0 & 17.6 & 38.7 & 19.7 & 31.8 \\
\midrule
Kanade 12.5Hz & 0.2 & \bfseries 78.8 & 69.6 & 44.1 & \bfseries 6.6 & 13.7 \\
Kanade 25Hz & 0.3 & 78.6 & 71.0 & 36.2 & 7.0 & 11.8 \\
\bottomrule
\end{tabular}
\end{table}

\textbf{Downstream speaker discrimination}\;
We also probe speaker information in different parts of disentangled codecs' representations (\textit{Content} only, \textit{Global} only and \textit{Both}) by training downstream discriminative models. Following \citet{jung2022pushing}, we train ECAPA-TDNN~\citep{desplanques2020ecapa} with AAM-softmax loss~\citep{deng2019arcface} on representations extracted from VoxCeleb1~\citep{nagrani20voxceleb}. We evaluate on speaker identification (SID) and automatic speaker verification (ASV) and report accuracy (Acc) and EER, respectively. Appendix~\ref{appendix:downstream} contains further details.

Results are shown in Table~\ref{table:disentangle-downstream}. 
We observe FACodec and TiCodec distribute most speaker information in their content tokens, which is consistent with the timbre transfer failure shown in the VC results. BiCodec exhibits a degree of separation, but it is relatively harder to access speaker information in its global representation. Kanade achieves the best speaker discrimination, confirming its SOTA disentanglement of content and speaker identity.

\subsubsection{Downstream speech recognition}

\begin{table}[t]
\caption{\textbf{Speech Recognition Results (\%)}. Bold numbers represent the best performance among codecs. For context, SOTA metric from specialized models~\citep{rekesh2023fast} is included.}
\label{table:asr}
\scriptsize
\centering
\setlength{\tabcolsep}{4pt}
\begin{tabular}{
    m{6.5em} 
    >{\centering\arraybackslash}m{2.5em} 
    S[table-format=3.1,round-mode=places,round-precision=1,detect-weight]
    S[table-format=2.1,round-mode=places,round-precision=1,detect-weight]
    S[table-format=3.1,round-mode=places,round-precision=1,detect-weight]
    S[table-format=2.1,round-mode=places,round-precision=1,detect-weight]
}
\toprule
\multirow{2}{*}{\textbf{Model}} & \multirow{2}{*}{\makecell{\textbf{Token} \\ \textbf{Rate}}} & \multicolumn{2}{c}{\textbf{LibriSpeech test-clean}} & \multicolumn{2}{c}{\textbf{SwitchBoard}} \\
\cmidrule(lr){3-4} \cmidrule(lr){5-6}
& & {WER$\downarrow$} & {CER$\downarrow$} & {WER$\downarrow$} & {CER$\downarrow$} \\
\midrule
\rowcolor{gray!10} SOTA & \NA & 1.38 & \NA & \NA & \NA \\
\rowcolor{gray!10} Cont. 50Hz & \NA & 4.31 & 1.93 & \NA & \NA \\
\rowcolor{gray!10} KM 12.5Hz & 12.5 & 5.79 & 2.91 & 17.3 & 11.2 \\
\rowcolor{gray!10} KM 25Hz & 25 & 5.81 & 3.13 & 15.0 & 9.5 \\
\midrule
BiCodec & 50 & 100.12 & 71.42 & 108.8 & 78.4 \\
WavTokenizer & 40 & 18.09 & 10.34 & 67.2 & 46.8 \\
StableCodec & 25 & 11.84 & 6.32 & 45.0 & 30.2 \\
X-Codec 2 & 50 & 10.97 & 5.96 & 103.3 & 75.6 \\
Mimi & 100 & 10.4 & 5.5 & 30.6 & 20.1 \\
DualCodec & 100 & 9.8 & 5.3 & 28.2 & 18.1 \\
TiCodec & 300 & 9.4 & 4.8 & 29.1 & 18.9 \\
FACodec & 480 & 8.2 & 4.2 & 25.5 & 16.4 \\
ST & 400 & 8.2 & 4.2 & 29.4 & 19.1 \\
PAST & 400 & 7.9 & 3.9 & 28.9 & 18.8 \\
\midrule
Kanade 12.5Hz & 12.5 & 8.13 & 4.01 & 24.6 & 15.9 \\
Kanade 25Hz & 25 & \bfseries 7.12 & \bfseries 3.75 & \bfseries 18.6 & \bfseries 11.7 \\
\bottomrule
\end{tabular}
\end{table}

To measure the availability of lexical information in tokens, we train decoder-only transformer ASR models following \citet{huang2024repcodec}. The models are trained to predict SentencePiece~\citep{kudo2018sentencepiece} text tokens conditioned on speech tokens extracted from the training sets of LibriSpeech and the SwitchBoard dataset~\citep{godfrey1993switchboard} of telephone conversations (out-of-distribution), respectively. We use all the token layers, but exclude global representations. Multi-layer tokens are embedded separately and concatenated. 
The resulting models (as well as the TTS models described below) have 85M backbone parameters. Appendix~\ref{appendix:downstream} contains further details. 
Figure~\ref{fig:downstream-models} in the Appendix illustrates our downstream models.

The results are shown in Table~\ref{table:asr}. Despite using much fewer tokens, Kanade 25Hz achieves the lowest WER (7.1\%, 18.6\%), drawing nearer to the performance of k-means tokens.
This indicates that our single stream captures rich and easily-accessible linguistic information and generalizes well. For metrics of phonetic information such as ABX and PNMI, see Appendix~\ref{appendix:phonetic}. For a correlation analysis between lexical and phonetic metrics, see Appendix~\ref{appendix:correlation}.

\subsubsection{Downstream text-to-speech}

\begin{table}
\caption{\textbf{Text-to-speech results.} Bold numbers are the best among codecs. For MUSHRA confidence intervals, see Tables~\ref{table:tts-quality-mushra} and \ref{table:tts-prosody-mushra}.}
\label{table:tts}
\centering
\scriptsize
\begin{tabular}{
    m{6.5em} 
    S[table-format=2.1,round-mode=places,round-precision=1,detect-weight]
    S[table-format=1.2,round-mode=places,round-precision=2,detect-weight]
    S[table-format=1.2,round-mode=places,round-precision=2,detect-weight]
    S[table-format=2.1,round-mode=places,round-precision=1,detect-weight]
    S[table-format=2.1,round-mode=places,round-precision=1,detect-weight]
    S[table-format=2.1,round-mode=places,round-precision=1,detect-weight]
    S[table-format=1.2,round-mode=places,round-precision=2,detect-weight]
}
\toprule
\multirow{2}{*}{\textbf{Model}} & 
\multicolumn{5}{c}{\textbf{LibriTTS test-clean}} & 
\multicolumn{2}{c}{\textbf{Seed-TTS-eval}} \\
\cmidrule(lr){2-6} \cmidrule(lr){7-8}
& {WER$\downarrow$} & {SIM$\uparrow$} & {UTMOS$\uparrow$} & {Quality$\uparrow$} & {Prosody$\uparrow$} & {WER$\downarrow$} & {SIM$\uparrow$} \\
\midrule
\rowcolor{gray!10} Ground Truth & 2.3 & \NA & 4.126 & 74.9 & 80.9 & 1.9 & \NA \\
\rowcolor{gray!10} CosyVoice 2 & 1.8 & 0.955 & 4.415 & 77.1 & 83.0 & 2.1 & 0.656\\
\rowcolor{gray!10} KM 12.5Hz & 4.6 & 0.947 & 3.961 & 72.0 & 67.0 & 5.4 & 0.419\\
\rowcolor{gray!10} KM 25Hz & 4.3 & 0.949 & 4.053 & 74.9 & 75.9 & 4.9 & 0.420\\
\midrule
WavTokenizer & 13.9 & 0.920 & 3.757 & 74.5 & 77.0 & 15.6 & 0.275\\
TiCodec & 11.5 & 0.939 & 3.857 & 73.8 & 72.9 & 12.9 & 0.325\\
DualCodec & 10.0 & \bfseries 0.957 & 3.675 & 73.0 & 80.0 & 5.5 & 0.343\\
ST & 9.7 & 0.953 & 3.953 & 75.0 & 79.0 & 11.2 & 0.352\\
StableCodec & 9.0 & 0.907 & 3.777 & 71.0 & 66.0 & 10.9 & 0.226\\
PAST & 8.0 & 0.954 & 4.139 & 74.9 & 78.4 & 9.0 & 0.354\\
BiCodec & 7.8 & 0.953 & 4.121 & 73.8 & 78.9 & 7.5 & 0.456\\
Mimi & 6.6 & 0.946 & 3.483 & 74.9 & 73.9 & 6.0 & 0.316\\
X-Codec 2 & 6.5 & 0.952 & \bfseries 4.205 & 72.0 & 78.0 & 7.2 & 0.347\\
\midrule
Kanade 12.5Hz & 5.9 & 0.952 & 4.133 & \bfseries 77.1 & 77.9 & 5.7 & 0.468\\
Kanade 25Hz & \bfseries 4.2 & 0.954 & 4.181 & 73.0 & \bfseries 81.0 & \bfseries 4.0 & \bfseries 0.480\\
\bottomrule
\end{tabular}
\end{table}

To test text-conditioned generative modeling, we train a decoder-only transformer phoneme-based TTS model for each tokenizer on the LibriTTS training sets.
Following \citet{borsos2023audiolm}, multi-layer tokens are flattened by interleaving them in a time-first order with a combined vocabulary\footnote{While some works prefer hierarchical modeling for efficiency~\citep{chen2025neural, defossez2024moshi}, we choose to standardize on flattening to allow simple and fair comparison among different tokenizers. \citet{copet2023simple} shows that autoregressive modeling on flattened tokens produces the highest quality results.}.
Following \citet{du2024cosyvoicescalablemultilingualzeroshot}, speaker identity is conditioned by prepending the input with WavLM-SV speaker embeddings from the reference. Global embeddings for synthesis after TTS modeling are also extracted from the reference. 
Training details are in Appendix~\ref{appendix:downstream}.
We randomly select 1,000 samples (4-10 seconds) from LibriTTS \texttt{test-clean} and condition each with 3 reference samples from the same speaker.
We also report Seed-TTS-eval~\citep{anastassiou2024seed} results for comparison with other works.
Quality and prosody are evaluated using MUSHRA-like listening tests (see Appendix~\ref{appendix:subjective}).

The results are shown in Table~\ref{table:tts}. On both test sets, Kanade achieves SOTA WER (4.2\%, 4.0\%) among codecs and SSL k-means tokens with excellent quality. This finding aligns with the ASR metrics discussed earlier: the stronger lexical availability in our content tokens provides more effective text alignment for downstream tasks. Also, Kanade achieves the best prosody naturalness (81.0) among codecs and k-means tokens, indicating that its rich prosodic features enable downstream models to generate expressive speech.

\subsubsection{Pure spoken language modeling}

\begin{table}[tb]
\caption{\textbf{Spoken language modeling results (\%).} Chance level is 50\%.}
\label{table:slm}
\scriptsize
\centering
\begin{tabular}{
    m{6.5em} 
    S[table-format=2.1]
    S[table-format=5]
    S[table-format=2.1,round-mode=places,round-precision=1,detect-weight]
    S[table-format=2.1,round-mode=places,round-precision=1,detect-weight]
    S[table-format=2.1,round-mode=places,round-precision=1,detect-weight]
    S[table-format=2.1,round-mode=places,round-precision=1,detect-weight]
}
\toprule
\textbf{Model} & \textbf{Token rate} & \textbf{Vocab. size} & \textbf{sWUGGY}$\uparrow$ & \textbf{sBLIMP}$\uparrow$ & \textbf{sSC}$\uparrow$ & \textbf{tSC}$\uparrow$ \\
\midrule
KM 12.5Hz & 12.5 & 12800 & 75.8 & \bfseries 57.5 & 51.8 & 66.7 \\
KM 25Hz & 25 & 12800 & 68.1 & 53.5 & 51.1 & 63.5 \\
\midrule
ST & 50 & 1024 & 75.8 & 54.9 & 52.0 & 64.4 \\
PAST & 50 & 1024 & 76.8 & 53.6 & 51.8 & 59.5 \\
Mimi & 12.5 & 2048 & \bfseries 77.6 & 56.1 & 52.0 & \bfseries 67.8 \\
\midrule
Kanade 12.5Hz & 12.5 & 12800 & 76.6 & 55.2 & \bfseries 52.1 & 65.3 \\
Kanade 25Hz & 25 & 12800 & 69.7 & 52.4 & 51.3 & 60.0 \\
\bottomrule
\end{tabular}
\end{table}

To test Kanade's suitability for pure spoken language modeling, we use the Slam~\citep{maimon2025slamming} recipe to train a warm-start SLM based on Qwen-2.5-0.5B on one epoch of LibriLight~\citep{kahn2020librilight}. We evaluate in-vocabulary sWUGGY, sBLIMP~\citep{dunbar21zero}, sStoryCloze (sSC), and tStoryCloze (tSC)~\citep{hassid2023textually}, all of which measure accuracy in assigning higher probability to linguistically plausible inputs. 
Since we keep constant the SLM architecture, these metrics indirectly measure whether a tokenizer makes available the information necessary to learn higher-level linguistic structure. Due to limited resources, only baselines that performed well in preliminary testing (see Appendix~\ref{appendix:slm}) are included here. Following \citet{defossez2024moshi}, for multi-layer codecs, only the first RVQ layer is used.

The results in Table~\ref{table:slm} show that Kanade 12.5Hz matches the performance of k-means and hybrid codec tokens. 
Combined with the previous results that Kanade provides richer linguistic information and high-quality synthesis, this demonstrates the potential of single-layer codecs for strong spoken language modeling without giving up the benefits of reconstruction-oriented codecs.

\subsection{Ablation studies}

We conduct an extensive set of ablation studies to verify the effectiveness of our design choices. Some results are shown here in Table~\ref{table:ablation-main}. See Appendix~\ref{appendix:more-ablation} for SSL layer selection, SSL encoder, model backbone, and other minor designs.

\begin{description}[leftmargin=0.5em, nolistsep, topsep=0pt, partopsep=0pt]
    \item[Dual-branch design] We train a model without a global branch, using only content tokens to reconstruct both SSL features and a mel spectrogram. This model shows heavy degradation on every metric. Despite its simplicity, the dual-branch disentangling design is indispensable, separating constant acoustic information and allowing the content branch to focus on rich linguistic content.

    \item[SSL feature reconstruction loss] In the model trained without SSL feature reconstruction loss, reconstruction and downstream ASR WERs are significantly higher. This suggests that the SSL feature reconstruction loss considerably encourages the content branch to encode linguistic information, compensating for the fact that the mel spectrogram loss is less sensitive to phonetic contrasts.

    \item[End-to-end training] In this setting, we first train the content FSQ-VAE with only SSL feature reconstruction loss, freeze it, and then train the other components with only mel spectrogram reconstruction loss. This is similar to the approach used by RepCodec~\citep{huang2024repcodec}. While this 2-stage method has a slightly lower WER, the speech quality, in particular prosody, degrades. This demonstrates that end-to-end training with dual objectives can extract more prosody without losing much lexical information.

    \item[FSQ] We replace FSQ with ordinary VQ~\citep{vandenoord2017neural}, using exponential moving average (EMA) codebook (decay 0.8), k-means initialization, and random restart for dead codes~\citep{dhariwal2020jukebox}. The results show a serious degradation on nearly every metric, especially linguistic ones (WER, F0Corr). This observation aligns with findings reported by \citet{mentzer2024finite}. FSQ yields better results and removes the need to tune extra hyperparameters.
\end{description}

\begin{table}[tb]
\caption{\textbf{Ablation results.} Based on Kanade 12.5Hz without post-training.}
\label{table:ablation-main}
\scriptsize
\centering
\setlength{\tabcolsep}{1pt}
\begin{tabular}{
    m{7.3em} 
    S[table-format=2.1,round-mode=places,round-precision=1,detect-weight]
    S[table-format=2.1,round-mode=places,round-precision=1,detect-weight]
    S[table-format=1.2,round-mode=places,round-precision=2,detect-weight]
    S[table-format=1.2,round-mode=places,round-precision=2,detect-weight]
    S[table-format=1.2,round-mode=places,round-precision=2,detect-weight]
    S[table-format=2.1,round-mode=places,round-precision=1,detect-weight]
}
\toprule
\multirow{2}{*}{\textbf{Model}} & 
\multicolumn{5}{c}{\textbf{Reconstruction}} & 
\multicolumn{1}{c}{\textbf{Downstream}} \\
\cmidrule(lr){2-6} \cmidrule(lr){7-7}
& {WER$\downarrow$} & {MUSHRA$\uparrow$} & {UTMOS$\uparrow$} & {SIM$\uparrow$} & {F0Corr$\uparrow$} & {WER$\downarrow$} \\
\midrule
Kanade 12.5Hz & 3.5 & 69.0 & 4.100 & 0.963 & 0.838 & 8.1 \\
\midrule
w/o Dual-branch & 6.1 & 24.0 & 2.925 & 0.882 & 0.656 & 10.4 \\
w/o Feature recon. & 8.0 & 68.5 & 4.083 & 0.958 & 0.839 & 14.9 \\
w/o End-to-end & 3.3 & 60.7 & 3.974 & 0.959 & 0.758 & 7.7 \\
w/o FSQ & 25.8 & 43.7 & 3.374 & 0.949 & 0.687 & 18.6 \\
\bottomrule
\end{tabular}
\end{table}

\section{Conclusion}
We introduced Kanade, a single-layer disentangled speech tokenizer that extracts compact tokens suitable for both generative and discriminative modeling.
It starts with SSL features that already represent speech in a structured latent space, enabling effective disentanglement and high data efficiency.
It uses reconstruction losses on both SSL and acoustic features, encouraging it to capture both phonetics and prosody.
A tight codebook-free quantization bottleneck, relieved of the need to encode acoustic constants by a global branch, tokenizes linguistic content without the need for auxiliary disentanglement methods.
Despite this simple design, Kanade achieves superior disentanglement, as measured by VC and speaker discrimination.
Moreover, it reaches best-in-class lexical availability as measured by downstream ASR and TTS, exhibits SLM performance competitive with SSL tokens, and maintains speech quality matching multi-layer codecs.
Kanade demonstrates that a single-layer codec can provide the desirable properties of both SSL tokens and codec tokens for spoken language modeling.

\clearpage

\section*{Impact Statement}
We recognize the potential for abuse using our models, especially when used for voice conversion. However, during GAN post-training the discriminator was very strong and we had to hobble it severely, indicating that the audio generated by our model can easily be detected. We acknowledge that the pretrained SSL encoder and our training data have biases and encourage anyone using our architecture to use debiasing techniques or train with a larger, more diverse dataset, as we also plan to do in the future.

\bibliography{mybib}
\bibliographystyle{custom}

\clearpage
\appendix

\begin{figure*}[t]
\centering
\includegraphics[width=0.8\linewidth]{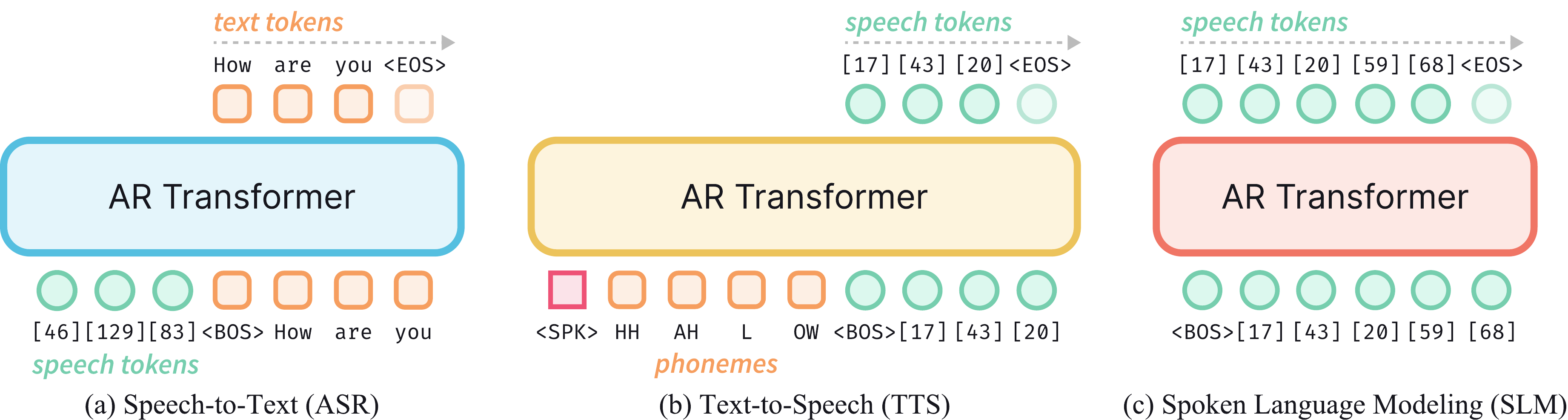}
\caption{\textbf{Downstream model architectures}}
\label{fig:downstream-models}
\end{figure*}

\section{Detailed comparison of other works}
\label{appendix:model-comparison}
\newcommand{\cmark}{\ding{51}}
\newcommand{\xmark}{\ding{55}}

\begin{table*}[tb]
\caption{\textbf{Comparison of partial recent speech codecs.} Mainly focus on disentangled codecs. Open-source status is as of January 2026.}
\label{table:model-comparison}
\scriptsize
\centering
\setlength{\tabcolsep}{4pt}
\begin{tabular}{lcccccc}
\toprule
\textbf{Model} & \textbf{Disentangled} & \textbf{Auxiliary Methods} & \makecell{\textbf{Use SSL}\\ \textbf{Features}} & \makecell{\textbf{Single-layer}\\ \textbf{Tokens}} & \makecell{\textbf{SSL Input}\\ \textbf{Only}} & \textbf{Open-source} \\
\midrule
PAST~\citep{har2025past} & \xmark & \NA & \xmark & \xmark & \NA & \cmark \\
StableCodec~\citep{parker2024scaling} & \xmark & \NA & \xmark & \cmark & \NA & \cmark \\
WavTokenizer~\citep{ji2024wavtokenizer} & \xmark & \NA & \xmark & \cmark & \NA & \cmark \\
SpeechTokenizer~\citep{zhang2024speechtokenizer} & \xmark & \NA & \cmark & \xmark & \xmark & \cmark \\
Mimi~\citep{defossez2024moshi} & \xmark & \NA & \cmark & \xmark & \xmark & \cmark \\
DualCodec~\citep{li25dualcodec} & \xmark & \NA & \cmark & \xmark & \xmark & \cmark \\
X-Codec 2~\citep{ye2025llasa} & \xmark & \NA & \cmark & \cmark & \xmark & \cmark \\
\midrule
FACodec~\citep{ju2024naturalspeech} & \cmark & \makecell{Supervision,\\Gradient Reversal} & \xmark & \xmark & \NA & \cmark \\
\addlinespace
Disen-TF-Codec~\citep{jiang2023disentangled} & \cmark & Instance Normalization & \xmark & \xmark & \NA & \xmark \\
PromptCodec~\citep{pan2024promptcodec} & \cmark & Structure Similarity Loss & \xmark & \xmark & \NA & \xmark \\
TiCodec~\citep{ren2024fewer} & \cmark & Invariance Learning & \xmark & \cmark & \NA & \cmark \\
Single-Codec~\citep{li2024single} & \cmark & Invariance Learning & \xmark & \cmark & \NA & \xmark \\
\addlinespace
SSVC~\citep{martin2024enhancing} & \cmark & \makecell{Contrastive Loss, \\Gradient Reversal} & \cmark & \xmark & \cmark & \xmark \\
\addlinespace
USC~\citep{vecino2025universal} & \cmark & \makecell{Contrastive Loss, \\Gradient Reversal, \\Differential Privacy} & \cmark & \xmark & \xmark & \xmark \\
\addlinespace
HAC~\citep{khurana2025factorized} & \cmark & Supervision & \cmark & \xmark & \xmark & \xmark \\
SoCodec~\citep{guo2024socodec} & \cmark & Perturbation & \cmark & \xmark & \xmark & \xmark \\
LSCodec~\citep{guo2024lscodec} & \cmark & Perturbation & \cmark & \cmark & \xmark & \xmark \\
FreeCodec~\citep{zheng2024freecodec} & \cmark & Perturbation & \cmark & \cmark & \xmark & \xmark \\
UniCodec~\citep{jiang2024universal} & \cmark & \xmark & \cmark & \xmark & \xmark & \xmark \\
\citet{aihara2025exploring} & \cmark & \xmark & \cmark & \xmark & \cmark & \xmark \\
HASRD~\citep{hussein2025hasrd} & \cmark & \xmark & \cmark & \xmark & \cmark & \xmark \\
BiCodec~\citep{wang2025spark} & \cmark & \xmark & \cmark & \cmark & \xmark & \cmark \\
\midrule
Kanade & \cmark & \xmark & \cmark & \cmark & \cmark & \cmark \\
\bottomrule
\end{tabular}
\end{table*}

Table~\ref{table:model-comparison} describes the tokenizers used as baselines in this paper, as well as all the disentangled codecs we are aware of. Most disentangled codecs use at least one auxiliary method to achieve disentanglement. However, in addition to BiCodec, which we include as a baseline, there are three other disentangling codecs that do not use auxiliary methods, including:
\citet{aihara2025exploring} and HASRD~\citep{hussein2025hasrd}, which are similar to our approach, operating directly on the SSL feature space, but using k-means residuals, resulting in multi-layer tokens; and
UniCodec~\citep{jiang2024universal}, which uses a triple-branch architecture and also produces multi-layer tokens. As these remain closed-source, we were unable to test them.

Beyond the scope of speech codecs, many studies explore speech tokenization for better language modeling, which we include here for completeness. dMel~\citep{bai2024dmel} demonstrates that training-free discretized mel spectrograms can enable effective downstream modeling, but produces long, multi-layer token sequences (80 layers, 40Hz) that requires architectural adaptations. 
The $\mathcal{S}^3$ tokenizer from CosyVoice~\citep{du2024cosyvoicescalablemultilingualzeroshot} extracts speech tokens by inserting a VQ layer into an ASR model and training with ASR objective. While effective, these tokens rely on costly large-scale supervision.
UniWav~\citep{liu2025uniwav} combines discriminative and generative objectives in a unified SSL framework, and the k-means tokens derived from their learned representations exhibit improved reconstruction quality.
However, this large-scale pretraining approach requires significant computational resources.

\section{Limitations and future work}
Since the SSL encoder we use is based on a bidirectional transformer, our tokens are not streamable, requiring audio chunking and limiting applicability in some scenarios. Nevertheless, we prototype chunk-based streaming in Appendix~\ref{appendix:chunked-encoding-decoding} to demonstrate practical usage. Since the effective receptive field of SSL encoders is limited~\citep{meng2025effective}, this can be solved by distilling a streamable encoder~\citep{choi25streaming} and modifying our architecture to a streaming design, as done by Mimi~\citep{defossez2024moshi}.

Our content tokens are produced at a constant rate, which may lead to information redundancy and reduce alignment with linguistic categories. We hope to adopt approaches pioneered by SyllableLM~\citep{baade2025syllablelm} and Sylber~\citep{cho2025sylber} to enable variable-rate tokenization, mitigating these issues.

Although we achieve excellent separation of dynamic content and acoustic constants, currently it is still not possible to further disentangle the content. It could be useful to further separate linguistic content into phonetic and prosodic features for better flexibility.

As shown by the GigaSpeech experiments (see Appendix~\ref{appendix:ood}), although Kanade generalizes to other domains to some extent, it is still sensitive to dynamic background noise such as music. This can likely be improved by training on a larger diverse data mixture or applying data augmentation. Some studies suggest fine-tuning the SSL encoder to enhance its noise robustness~\citep{gat2023augmentation, chang23spin, messica24nast}, which can be also applied to our encoder.
For more limitations regarding out-of-distribution data, see Appendix~\ref{appendix:ood}.

Since the focus of this paper is to improve linguistic availability in discrete speech tokens, we did not experiment with any vocoding settings other than targeting a mel spectrogram and using Vocos to generate a waveform. To improve audio quality, we might consider training a more advanced decoder such as flow-matching model on Kanade tokens.

\section{Additional ablation studies}
\label{appendix:more-ablation}
\subsection{Content branch}
\label{appendix:ablation-content}

\begin{table}[t]
\caption{\textbf{Content branch ablation results}}
\label{table:ablation-content}
\scriptsize
\centering
\setlength{\tabcolsep}{1pt}
\begin{tabular}{
    m{11em} 
    S[table-format=2.1,round-mode=places,round-precision=1,detect-weight]
    S[table-format=1.2,round-mode=places,round-precision=2,detect-weight]
    S[table-format=1.2,round-mode=places,round-precision=2,detect-weight]
    S[table-format=1.2,round-mode=places,round-precision=2,detect-weight]
    S[table-format=2.1,round-mode=places,round-precision=1,detect-weight]
}
\toprule
\multirow{2}{*}{\textbf{Model}} & 
\multicolumn{4}{c}{\textbf{Reconstruction}} & 
\multicolumn{1}{c}{\textbf{Downstream}} \\
\cmidrule(lr){2-5} \cmidrule(lr){6-6}
& {WER$\downarrow$} & {UTMOS$\uparrow$} & {SIM$\uparrow$} & {F0Corr$\uparrow$} & {WER$\downarrow$} \\
\midrule
Kanade 12.5Hz & 3.5 & 4.100 & 0.963 & 0.838 & 8.1 \\
\midrule
Token rate 6.25Hz & 14.0 & 3.547 & 0.946 & 0.646 & 15.8 \\
Codebook size 3125 & 4.9 & 4.045 & 0.958 & 0.788 & 10.0 \\
\midrule
Layer 6 & 4.2 & 4.088 & 0.963 & 0.817 & 12.5 \\
Layer 9 & 3.5 & 4.067 & 0.960 & 0.809 & 7.5 \\
Layer 12 & 3.5 & 4.044 & 0.960 & 0.800 & 7.8 \\
Layer 9+12 & 3.6 & 4.075 & 0.960 & 0.808 & 7.4 \\
Layer 1–12 weighted-sum & 3.5 & 4.074 & 0.960 & 0.802 & 8.7 \\
\bottomrule
\end{tabular}
\end{table}

Results of ablation on the content branch are shown in Table~\ref{table:ablation-content}.

We tried decreasing the \textbf{token rate} and \textbf{effective codebook size}. When the token rate is halved (6.25Hz), the linguistic content and speech quality is unacceptable. On the other hand, the codebook size has more moderate effect on information capacity, since the bitrate decreases logarithmically with codebook size. In the model with 3,125 codes ($\sim 1/4$ of the original codebook size, 145bps), WER and F0Corr mildly degrade.

We also study \textbf{SSL feature layer selection} for the content branch input. We observe a pattern consistent with \citet{pasad2023comparative}: shallow layers provide more acoustic information that benefits audio quality and prosody preservation; deep layers offer more phonetic information. We find the 9th layer ($3/4$ the way through) is a good balance point. \citet{zhang2025vevo} observed a similar result for HuBERT-large. Adding layer 6 to layer 9 improves speech quality and prosody, without losing much lexical availability (+0.6\% downstream WER), so we stick to this combination. 
We also experimented with a learnable weighted-sum of all layers, with suboptimal results. Interestingly this model distributes over 80\% of the weight to the deepest layer.
We did not perform full sweep on the SSL layers because prior works already did that and we want to focus on tokenizer architecture design in this work.

\subsection{Global branch}
\label{appendix:ablation-global}

\begin{table}[t]
\caption{\textbf{Global branch ablation results}}
\label{table:ablation-global}
\scriptsize
\centering
\setlength{\tabcolsep}{0.5pt}
\begin{tabular}{
    m{10.5em} 
    S[table-format=2.1,round-mode=places,round-precision=1,detect-weight]
    S[table-format=1.2,round-mode=places,round-precision=2,detect-weight]
    S[table-format=1.2,round-mode=places,round-precision=2,detect-weight]
    S[table-format=1.2,round-mode=places,round-precision=2,detect-weight]
    S[table-format=2.1,round-mode=places,round-precision=1,detect-weight]
    S[table-format=2.1,round-mode=places,round-precision=1,detect-weight]
}
\toprule
\multirow{2}{*}{\textbf{Model}} & 
\multicolumn{4}{c}{\textbf{Reconstruction}} & 
\multicolumn{2}{c}{\textbf{Downstream}} \\
\cmidrule(lr){2-5} \cmidrule(lr){6-7}
& {WER$\downarrow$} & {UTMOS$\uparrow$} & {SIM$\uparrow$} & {F0Corr$\uparrow$} & {SID Acc$\uparrow$} & {ASV EER$\downarrow$} \\
\midrule
Kanade 12.5Hz & 3.5 & 4.100 & 0.963 & 0.838 & 69.6 & 13.7 \\
\midrule
Layer 6+9 & 3.6 & 4.056 & 0.942 & 0.788 & 71.7 & 13.8 \\
Layer 1–4 weighted-sum & 3.7 & 4.093 & 0.966 & 0.821 & 75.4 & 10.9 \\
Mel & 3.6 & 3.805 & 0.931 & 0.807 & 46.3 & 20.0 \\
\midrule
Avg pooling & 3.7 & 4.104 & 0.962 & 0.810 & 70.3 & 12.6 \\
Conditioning: full decoder & 3.8 & 4.094 & 0.963 & 0.822 & 70.9 & 11.8 \\
Conditioning: addition & 3.8 & 4.085 & 0.969 & 0.829 & 82.6 & 12.7 \\
\bottomrule
\end{tabular}
\end{table}

Results of ablation on the global branch are shown in Table~\ref{table:ablation-global}.

For \textbf{SSL feature layer selection}, we experiment with using the same combination of SSL layers as our content branch (layers 6 and 9) and observe that prosody metrics are worse.

In a model with a learnable weighted-sum of layers 1–4, we notice increased speaker recognition performance (75.4\% SID Acc) but slightly worse intelligibility. Other metrics remain similar. During training, we find the model distributes over 50\% of the weight to layer 1, indicating that the global branch prefers information from earlier layers. For simplicity, we stick to a combination of layers 1 and 2 for better intelligibility while maintaining reasonably high downstream performance.

We also experiment with using mel spectrograms as input for global branch instead of SSL features. This worsened all metrics. This indicates that SSL features provide more useful and structured information on speaker identity, benefiting both reconstruction and downstream performance. This result motivated us to build a tokenizer fully based on SSL features.

Moreover, we study the effect of \textbf{pooling and conditioning} in the global branch. Compared to average pooling, our main model with attentive statistical pooling~\citep{okabe2018attentive} has slightly better intelligibility and prosody. For conditioning mechanism ablation, we train (1) a variant where global embeddings apply adaLN-Zero~\citep{peebles2023scalable} conditioning to both the token module and mel module in our decoder instead of just the latter (noted as Conditioning: full decoder), and (2) a variant using simple addition instead of adaLN-Zero (noted as Conditioning: addition). Both of them exhibit slightly worse intelligibility and prosodic correlation, though the model with addition conditioning achieves remarkable SID accuracy (82.6\%). We stick to adaLN-Zero conditioning only mel module, as this seems to better preserve linguistic information.

\subsection{Architecture}
\label{appendix:ablation-arch}

\begin{table}[t]
\caption{\textbf{Backbone and SSL encoder ablation results}}
\label{table:ablation-arch}
\scriptsize
\centering
\setlength{\tabcolsep}{1pt}
\begin{tabular}{
    m{6.5em} 
    S[table-format=2.1,round-mode=places,round-precision=1,detect-weight]
    S[table-format=1.2,round-mode=places,round-precision=2,detect-weight]
    S[table-format=1.2,round-mode=places,round-precision=2,detect-weight]
    S[table-format=1.2,round-mode=places,round-precision=2,detect-weight]
    S[table-format=2.1,round-mode=places,round-precision=1,detect-weight]
    S[table-format=2.1,round-mode=places,round-precision=1,detect-weight]
    S[table-format=2.1,round-mode=places,round-precision=1,detect-weight]
}
\toprule
\multirow{2}{*}{\textbf{Model}} & 
\multicolumn{4}{c}{\textbf{Reconstruction}} & 
\multicolumn{3}{c}{\textbf{Downstream}} \\
\cmidrule(lr){2-5} \cmidrule(lr){6-8}
& {WER$\downarrow$} & {UTMOS$\uparrow$} & {SIM$\uparrow$} & {F0Corr$\uparrow$} & {WER$\downarrow$} & {SID Acc$\uparrow$} & {ASV EER$\downarrow$} \\
\midrule
Kanade 12.5Hz & 3.5 & 4.100 & 0.963 & 0.838 & 8.1 & 69.6 & 13.7 \\
\midrule
ConvNeXt & 4.0 & 4.044 & 0.962 & 0.823 & 8.9 & 73.1 & 10.0 \\
HuBERT & 3.7 & 4.093 & 0.960 & 0.819 & 9.2 & 65.7 & 10.5 \\
\bottomrule
\end{tabular}
\end{table}
\begin{table}[tb]
\caption{\textbf{GAN post-training ablation results}}
\label{table:ablation-gan}
\scriptsize
\centering
\begin{tabular}{
    m{6.5em} 
    S[table-format=2.1,round-mode=places,round-precision=1,detect-weight]
    S[table-format=2.1,round-mode=places,round-precision=1,detect-weight]
    S[table-format=1.2,round-mode=places,round-precision=2,detect-weight]
    S[table-format=1.2,round-mode=places,round-precision=2,detect-weight]
    S[table-format=1.2,round-mode=places,round-precision=2,detect-weight]
}
\toprule
\multirow{2}{*}{\textbf{Model}} & 
\multicolumn{5}{c}{\textbf{Reconstruction}} \\
\cmidrule(lr){2-6}
& {WER$\downarrow$} & {MUSHRA$\uparrow$} & {UTMOS$\uparrow$} & {SIM$\uparrow$} & {F0Corr$\uparrow$} \\
\midrule
Kanade 12.5Hz & 3.35 & 74.6 & 4.168 & 0.966 & 0.847 \\
\quad w/o GAN & 3.46 & 69.0 & 4.100 & 0.963 & 0.838 \\
\midrule
Kanade 25Hz & 2.35 & 75.0 & 4.164 & 0.972 & 0.883 \\
\quad w/o GAN & 2.34 & 70.3 & 4.128 & 0.969 & 0.878 \\
\bottomrule
\end{tabular}
\end{table}

We train a model with all transformers replaced with ConvNeXt~\citep{liu2022convnet} backbones with a matching parameter count. The results are in Table~\ref{table:ablation-arch}. The model shows similar results except mildly worse linguistic content metrics (+0.5\% reconstruction WER and +0.8\% downstream WER). This indicates that the stronger sequence modeling ability of transformers can help the model better preserve and surface linguistic information.

We also try replacing WavLM Base+ with HuBERT-base, which shows similar results in Table~\ref{table:ablation-arch}. This validates the effectiveness of our method across SSL models.

Table~\ref{table:ablation-gan} shows reconstruction results without GAN post-training. Based on these ablations, post-training slightly improves audio quality (higher MUSHRA, UTMOS) without heavily affecting other metrics.

\section{Analysis}
\subsection{Prosodic information probing}
\label{appendix:prosody}

\begin{table}[tb]
\caption{\textbf{Probing results on fundamental frequency (F0).} For multi-layer codecs, only the linguistically dense token layer (usually the first RVQ layer) is used.}
\label{table:f0-probe}
\scriptsize
\centering
\setlength{\tabcolsep}{4pt}
\begin{tabular}{
    m{6.5em} 
    S[table-format=1.2,round-mode=places,round-precision=2,detect-weight]
    S[table-format=1.2,round-mode=places,round-precision=2,detect-weight]
}
\toprule
\textbf{Model} & \textbf{Corr}$\uparrow$ & \textbf{RMSE}$\downarrow$ \\
\midrule
KM 12.5Hz       & 0.504 & 0.856 \\
KM 25Hz         & 0.529 & 0.839 \\
\midrule
DualCodec       & \bfseries 0.781 & \bfseries 0.619 \\
WavTokenizer    & \bfseries 0.778 & 0.627 \\
FACodec         & 0.638 & 0.762 \\
ST              & 0.566 & 0.817 \\
X-Codec 2       & 0.551 & 0.828 \\
BiCodec         & 0.495 & 0.860 \\
Mimi            & 0.464 & 0.876 \\
StableCodec     & 0.349 & 0.927 \\
PAST         & 0.538 & 0.833 \\
TiCodec      & 0.676 & 0.729 \\
\midrule
Kanade 12.5Hz     & 0.676 & 0.729 \\
Kanade 25Hz       & 0.753 & 0.651 \\
\bottomrule
\end{tabular}
\end{table}

To measure the availability of prosodic information within speech tokens, we conduct a probing analysis on fundamental frequency (F0), which humans perceive as pitch. We train a small 7M-parameter (2-layer 512-dim) bidirectional transformers with a linear head to predict log F0. The models are optimized with MSE loss for 50k steps, using AdamW~\citep{loshchilov2018decoupled} (learning rate 1e-3, $\beta_1=0.9, \beta_2=0.999$, weight decay 1e-2). We use LibriSpeech \texttt{train-clean-100} for training and \texttt{test-clean} for testing. F0 extraction settings match those in our reconstruction experiments (Section~\ref{section:reconstruction}). Since our main focus is to investigate the usefulness of different speech tokens for prosody modeling in speech LMs, we use tokens from the most linguistically-related layer (RVQ 1, or the first content layer in FACodec) for multi-layer codecs; we use the single-layer version of TiCodec. The log F0 values are normalized for each instance, as only relative pitch is linguistically relevant. We report Pearson correlation coefficient (Corr) and root mean squared error (RMSE).

\begin{figure}[tb]
\centering
\includegraphics[width=\linewidth,trim={0.5cm 0.5cm 0.5cm 0.5cm},clip]{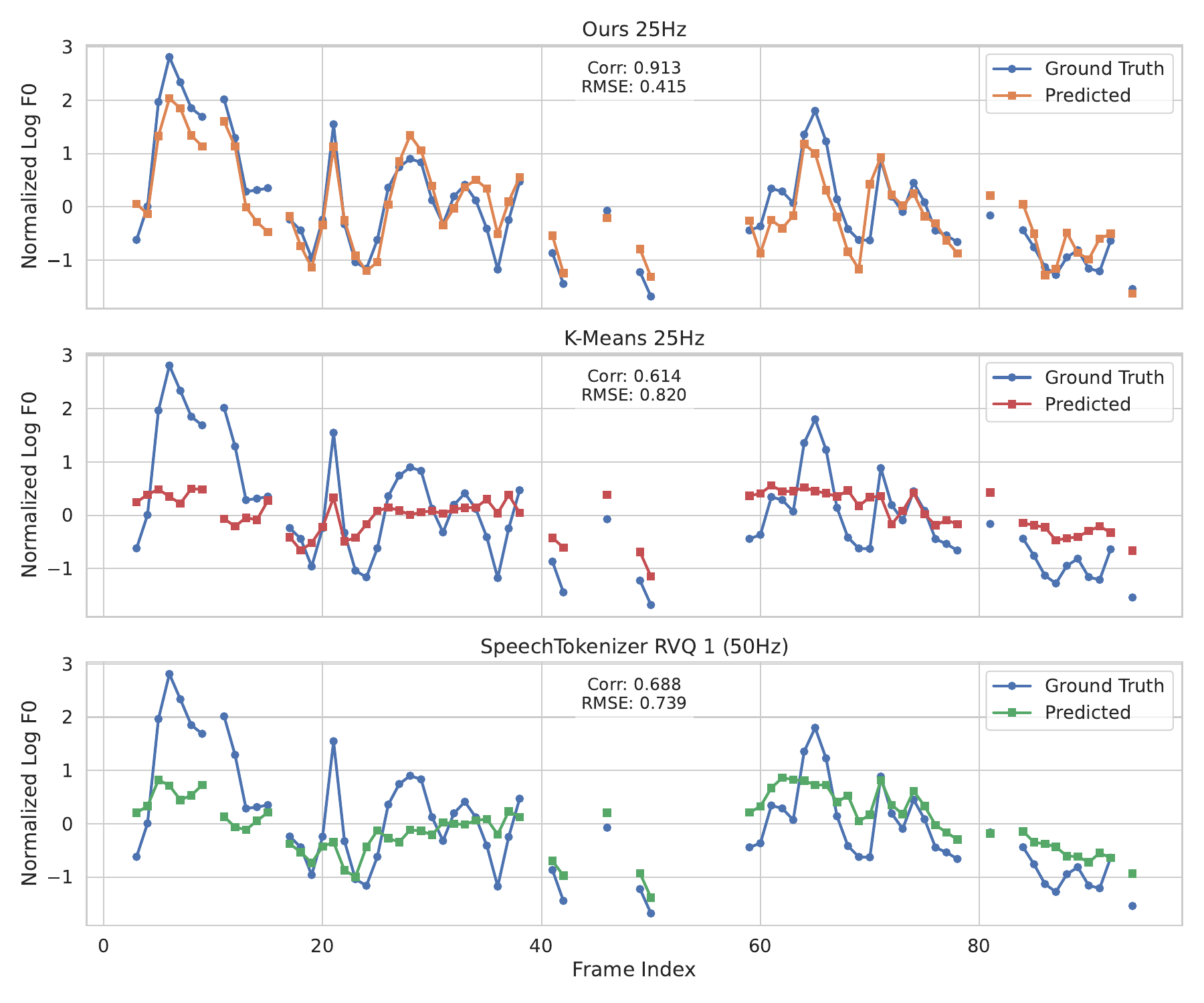}
\caption{\textbf{Comparison of F0 probing predictions.} The example is from LibriSpeech \texttt{1320-122617-29}.}
\label{fig:f0-comparison}
\end{figure}

Results are shown in Table~\ref{table:f0-probe}. Kanade models achieve better F0 probing performance than most of the baselines and k-means tokens (Kanade 25Hz 0.75 vs. KM 25Hz 0.53 on Corr). We display probing results for one sample in Figure~\ref{fig:f0-comparison}: predictions from our content tokens are more aligned with the ground truth than those from k-means or SpeechTokenizer. These results verify that our tokens make prosody easily accessible.

\subsection{Phonetic information analysis}
\label{appendix:phonetic}

\begin{figure}[tb]
\centering
\includegraphics[width=0.8\linewidth,trim={0.5cm 2.5cm 0.5cm 0.5cm},clip]{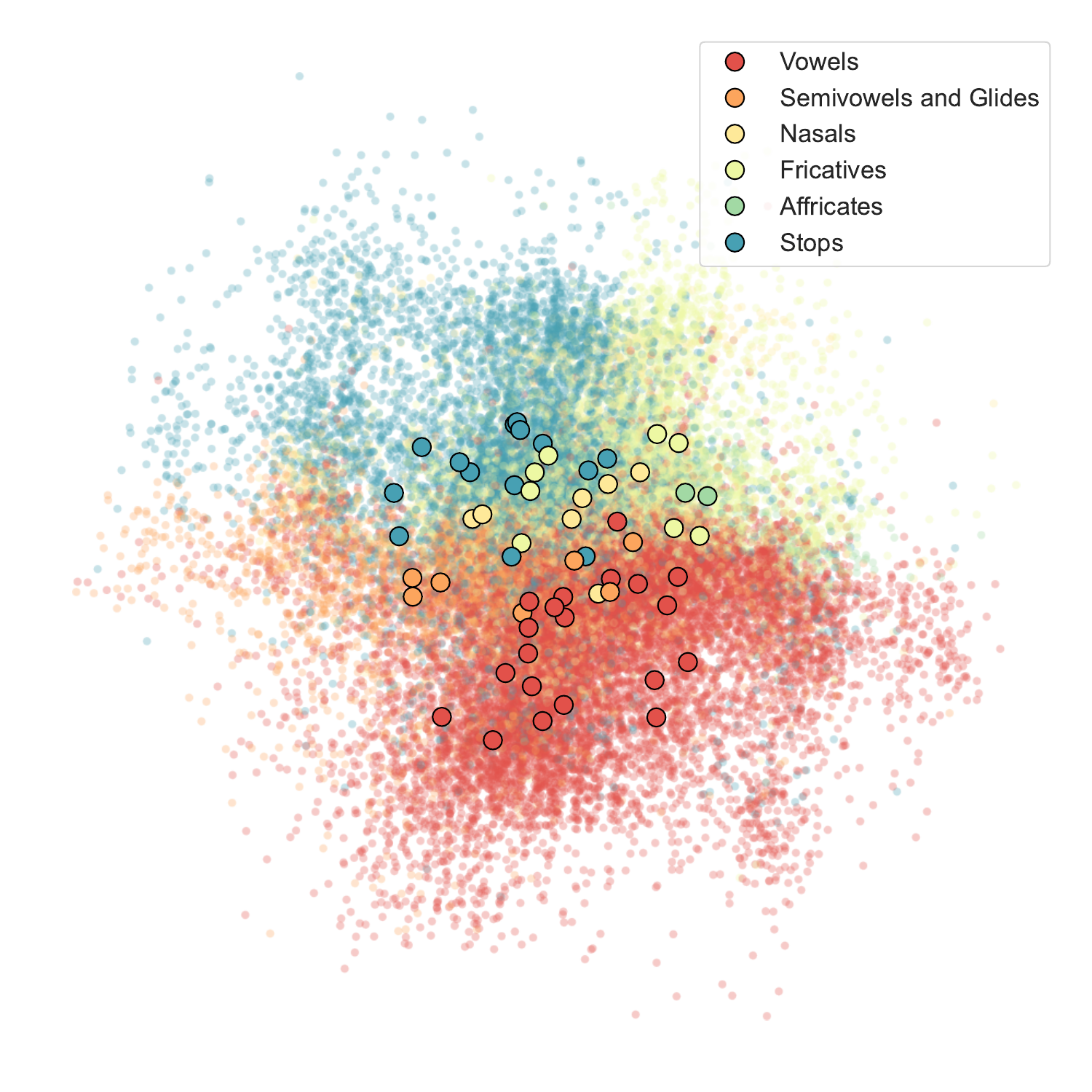}
\caption{\textbf{PCA visualization of our content embedding.} Points are colored by category. Larger markers represent per-phoneme average embeddings.}
\label{fig:phone-pca}
\end{figure}

First, we visualize the distribution of phones in the continuous content embedding space of our 12.5Hz model. We encode the TIMIT dataset~\citep{garofolo1993timit} using the content encoder (768-dim) and find the average embedding for each phoneme. We perform Principal Component Analysis (PCA) with two components on these average embeddings, then project all the collected embeddings onto the learned PCA space. The result is shown in Figure~\ref{fig:phone-pca}. We observe a clear phonetic configuration of the embedding space.

To numerically evaluate the phonemic information in our content tokens, we measure \textbf{ABX} phoneme discriminability~\citep{schatz2013abx} and phone-normalized mutual information (\textbf{PNMI})~\citep{hsu2021hubert}. In the literature on speech representations, the phone/phoneme terminology is not well-respected. We use terms as used in the original definitions of these metrics. Technically, both of them measure phonemic information, but hierarchical clustering shows that SSL representations are mostly phonetic~\citep{niekerk2023rhythm}.

\textbf{ABX} measures the extent to which phonemic categories are localized in feature space. It starts with a minimal pair of triphones like ``bag" and ``beg". The model is presented with $A$, an instance of the first, $B$, an instance of the second, and $X$, another instance of one of the two triphones. A and B always come from the same speaker. X either comes from the same speaker (\textit{within}) or a different speaker (\textit{across}).

We choose a distance measure $d(x, y)$ and calculate both $d(X, A)$ and $d(X, B)$. In a well-configured embedding, $X$ should be closer to the sample from the same class. For example, if $A$ is an instance of ``bag'', $B$ is instance of ``beg'', and $X$ is another instance of ``bag'', then we expect $d(X, A) < d(X, B)$.
The ABX score is the error rate: lower ABX scores indicate better phonemic discriminability directly in the representation space. We evaluate ABX on Libri-light~\citep{kahn2020librilight} \texttt{test-clean}, using the fastabx library~\citep{poli2025fastabx}. We use cosine similarity as the distance measure following convention~\citep{dunbar21zero}.

\begin{table}[tb]
\caption{\textbf{Phonetic information metrics.} For multi-layer codecs, only the linguistically dense token layer (usually the first RVQ layer) is used.}
\label{table:phonetic-metrics}
\scriptsize
\centering
\setlength{\tabcolsep}{4pt}
\begin{tabular}{
    m{6.5em} 
    S[table-format=2.1,round-mode=places,round-precision=1,detect-weight]
    S[table-format=2.1,round-mode=places,round-precision=1,detect-weight]
    S[table-format=1.2,round-mode=places,round-precision=2,detect-weight]
}
\toprule
\multirow{2}{*}{\textbf{Model}} & 
\multicolumn{2}{c}{\textbf{ABX}$\downarrow$} & 
{\textbf{PNMI}$\uparrow$} \\
\cmidrule(lr){2-3}
& {within} & {across} & \\
\midrule
KM 12.5Hz & 4.4\% & 5.1\% & 0.792 \\
KM 25Hz & 3.5\% & \bfseries 4.2\% & 0.810 \\
\midrule
PAST & \bfseries 3.4\% & \bfseries 4.2\% & \bfseries 0.865  \\
ST & 3.6\% & 4.5\% & 0.685  \\
FACodec & 4.4\% & 5.9\% & 0.534  \\
Mimi & 6.6\% & 7.8\% & 0.630  \\
X-Codec 2 & 15.4\% & 22.4\% & 0.442  \\
DualCodec & 16.0\% & 19.1\% & 0.559  \\
StableCodec & 21.9\% & 25.0\% & 0.547 \\
TiCodec & 22.1\% & 27.1\% & 0.177  \\
BiCodec & 24.5\% & 34.3\% & 0.224  \\
WavTokenizer & 25.6\% & 31.5\% & 0.170  \\
\midrule
Kanade 12.5Hz & 22.7\% & 24.3\% & 0.582  \\
Kanade 25Hz & 19.0\% & 21.6\% & 0.487  \\
\bottomrule
\end{tabular}
\end{table}

\textbf{PNMI} calculates the mutual information between phones and tokens $I(\text{phones};\text{tokens})$, normalized by phone entropy $H(\text{phones})$. It measures the amount of uncertainty about the phone identity that is eliminated by observing the token. Higher PNMI score indicates stronger correspondence between tokens and phones. We evaluate on the TIMIT dataset~\citep{garofolo1993timit}.

The results are shown in Table~\ref{table:phonetic-metrics}. K-means tokens achieve the best performance on these metrics, indicating a strong relationship with phonetic categories. PAST and FACodec, which uses phoneme labels, as well as SpeechTokenizer and Mimi, which use knowledge distillation, exhibit comparable performance. X-Codec 2, DualCodec, and Kanade, which use VQ-VAE, perform similarly.
See Appendix~\ref{appendix:correlation} for detailed analysis.

\begin{figure*}[p]
\centering
\includegraphics[width=0.48\linewidth]{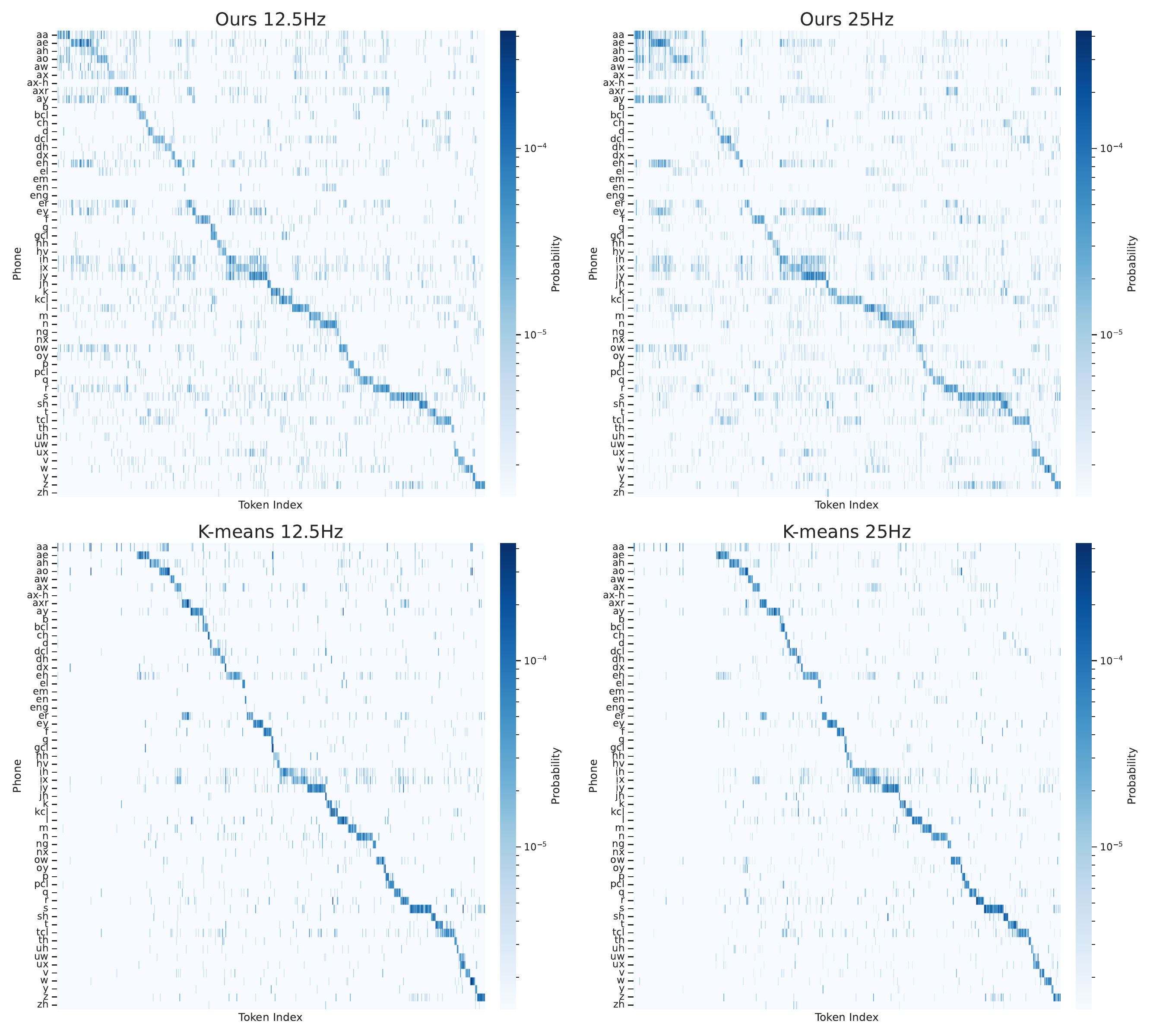}
\includegraphics[width=0.48\linewidth]{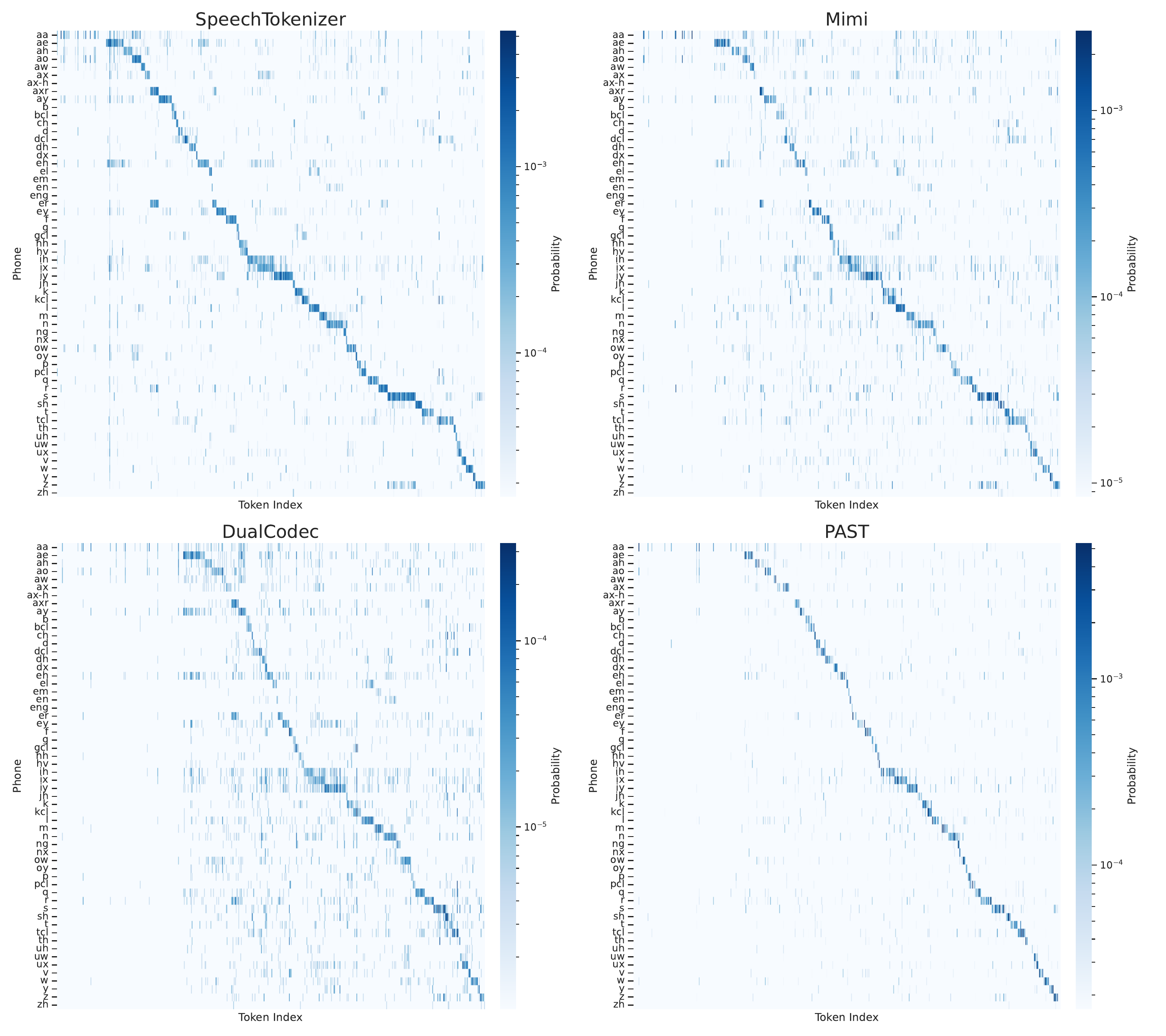}
\includegraphics[width=0.48\linewidth,align=t]{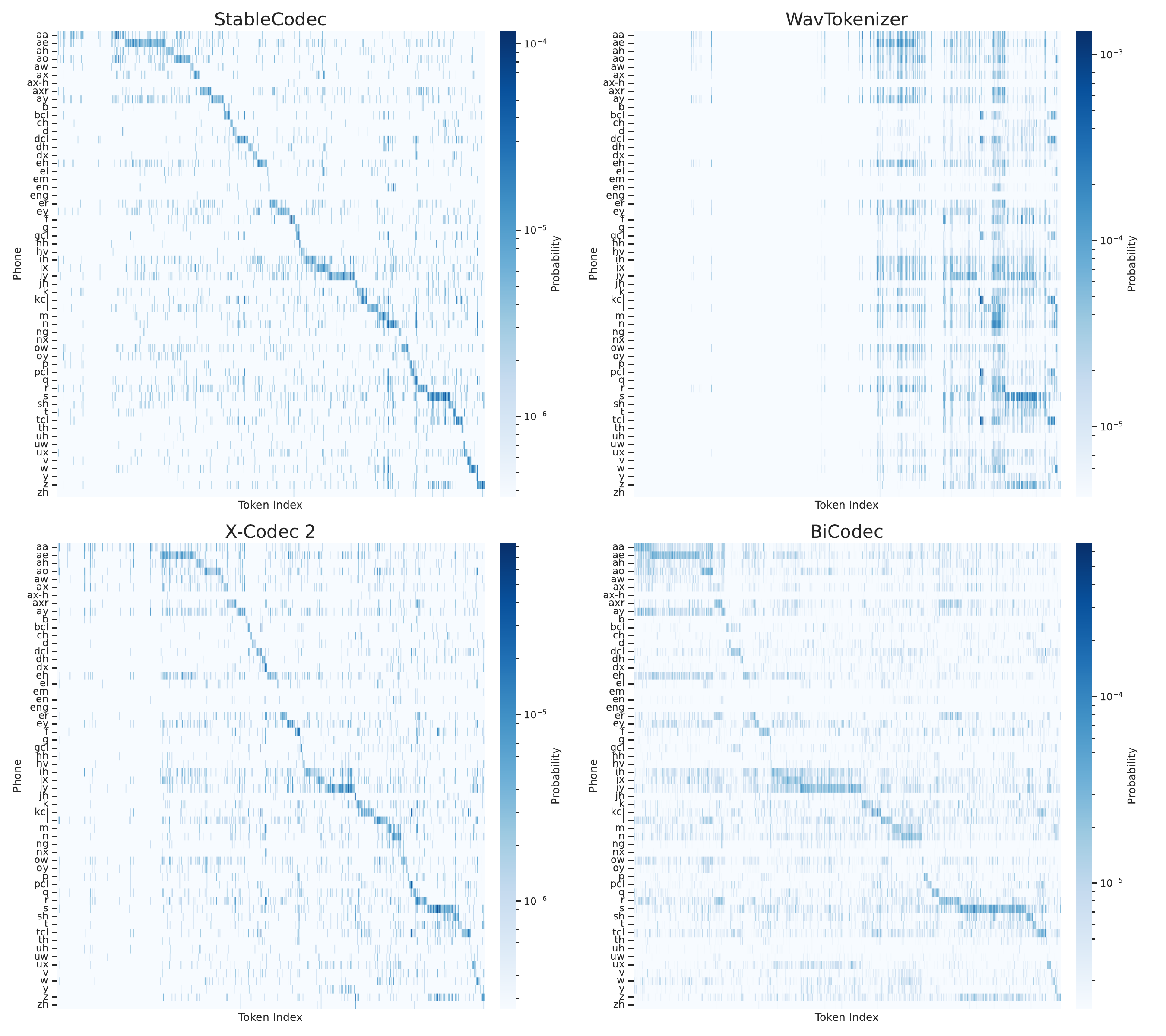}
\includegraphics[width=0.48\linewidth,align=t]{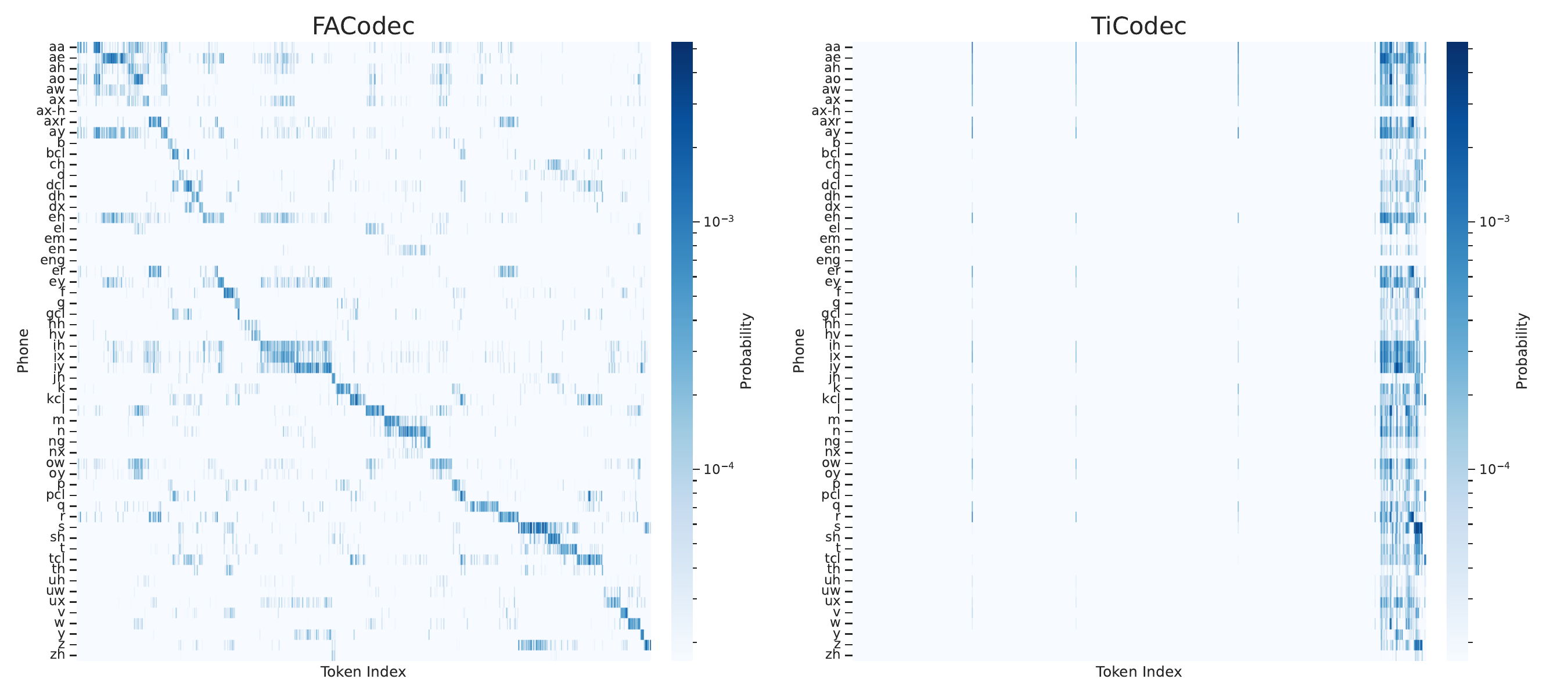}
\caption{\textbf{Joint probability distributions on speech tokens and TIMIT phones.} The token indices are sorted for better visualization. For multi-layer codecs, only the linguistically dense token layer (usually the first RVQ layer) is used.}
\label{fig:phone-matrices}
\end{figure*}

In Figure~\ref{fig:phone-matrices}, we visualize the relationship between speech tokens and phonemes. PNMI is a measure of the strength of this relationship. Though noisier than k-means tokens, ours show recognizable correspondence to TIMIT phonemes. Curiously, all tokenizers other than BiCodec, FACodec, and ours have a significant token space that is unrelated to encoding this information.

\begin{figure}[t]
\centering
\includegraphics[width=\linewidth]{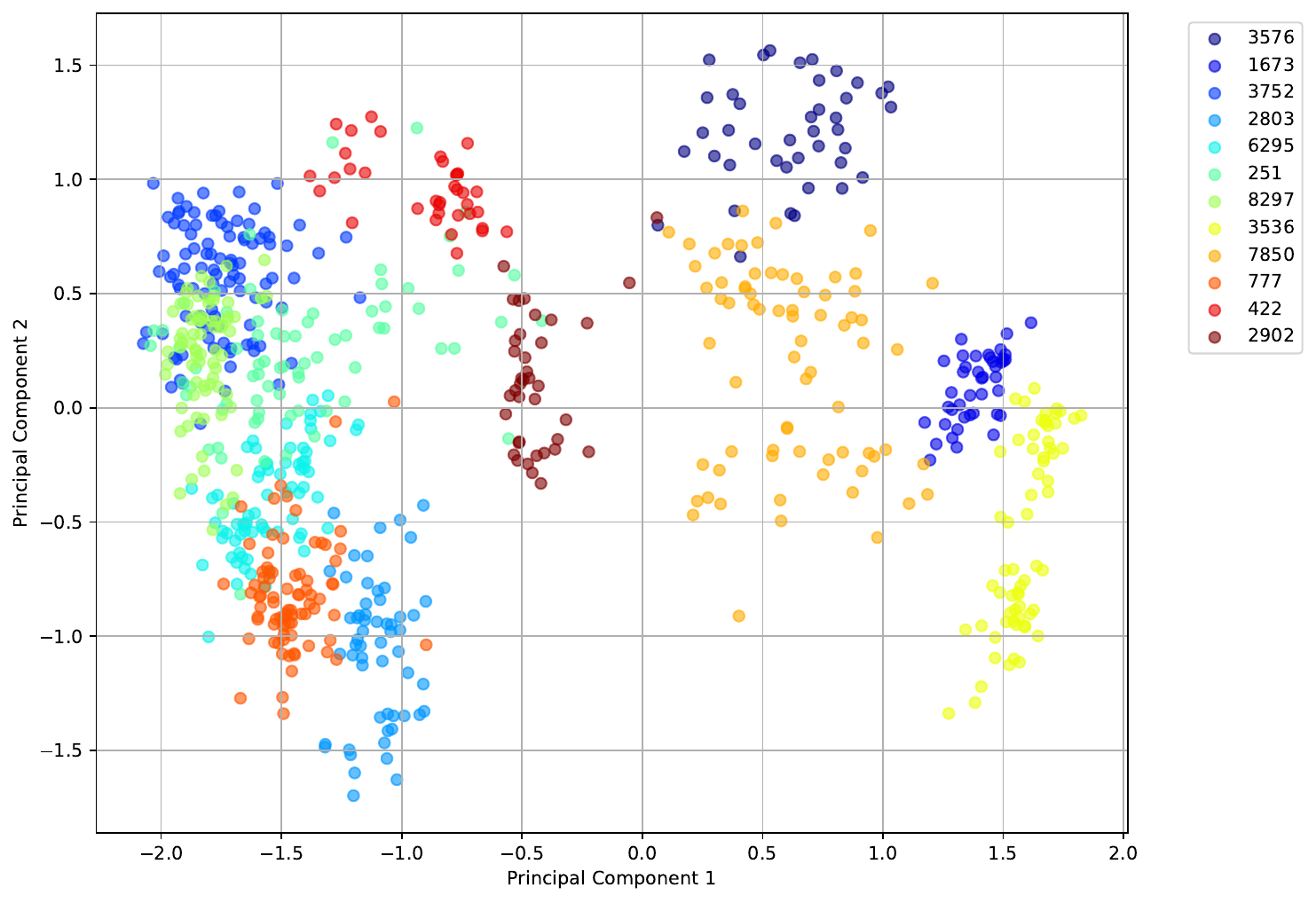}
\caption{\textbf{PCA of global embeddings.} Colored by LibriSpeech speaker ID.}
\label{fig:speaker_embeddings}
\end{figure}

\subsection{Global embedding PCA analysis}\label{appendix:global-embedding-pca}

We perform PCA on the global embeddings from LibriSpeech \texttt{dev-clean} and plot a subset of the utterances in Figure~\ref{fig:speaker_embeddings}.

To get a sense of what these components represent, we took utterances, tokenized them, and reconstructed them using a perturbed global embedding. Subjectively, the first principal component seems related to speaker gender. The second and third are harder to characterize without further analysis. Samples of these perturbations are available on the demo page\footnote{\url{https://frothywater.github.io/kanade-tokenizer/}}.

\subsection{Preliminary SLM results}
\label{appendix:slm}

Before training the SLMs described in main text, we trained weaker SLMs for each tokenizer using the training subset of LibriSpeech. For multi-layer tokenizers, tokens are extracted from the first RVQ layer (for FACodec, the first content layer), as those layers are meant to contain linguistic information for language modeling. We use the single-layer version of TiCodec. Each training sequence is randomly cropped to 20.48 seconds. An autoregressive transformer with 85M parameters (excluding embedding and output projection) is trained for 200k steps, with a batch size of 16. We use the last checkpoint for evaluation. Other transformer details are consistent with the descriptions in Appendix~\ref{appendix:downstream}.

Results are shown in Table~\ref{table:slm-preliminary}. Hybrid codecs (PAST, SpeechTokenizer, and Mimi) and k-means, both of which are phonetically dense perform the best. Kanade exposes more prosodic information (see Appendix~\ref{appendix:prosody}) in its one token stream, which may make learning more difficult, but as shown in the main text, using more powerful models can erase the gap.

\begin{table}[t]
\caption{\textbf{Preliminary SLM results (\%).} For multi-layer codecs, only the linguistically dense token layer (usually the first RVQ layer) is used.}
\label{table:slm-preliminary}
\scriptsize
\centering
\begin{tabular}{
    m{6.5em} 
    S[table-format=2.1]
    S[table-format=5]
    S[table-format=2.1,round-mode=places,round-precision=1,detect-weight]
    S[table-format=2.1,round-mode=places,round-precision=1,detect-weight]
}
\toprule
\textbf{Model} & \textbf{Token rate} & \textbf{Vocab. size} & \textbf{sWUGGY}$\uparrow$ & \textbf{sBLIMP}$\uparrow$ \\
\midrule
KM 12.5Hz & 12.5 & 12800 & 69.81 & \bfseries 54.04 \\
KM 25Hz & 25 & 12800 & 66.83 & 53.34 \\
\midrule
PAST & 50 & 1024 & \bfseries 75.03 & 52.30 \\
ST & 50 & 1024 & 70.97 & 52.14 \\
Mimi & 12.5 & 2048 & 68.31 & 53.69 \\
FACodec & 80 & 1024 & 57.24 & 50.13 \\
StableCodec & 25 & 46656 & 57.06 & 51.10 \\
DualCodec & 12.5 & 16384 & 56.52 & 50.15 \\
BiCodec & 50 & 8192 & 54.11 & 50.17 \\
WavTokenizer & 40 & 4096 & 52.67 & 50.81 \\
TiCodec & 75 & 1024 & 52.58 & 50.90 \\
X-Codec 2 & 50 & 65536 & 52.40 & 49.97 \\
\midrule
Kanade 12.5Hz & 12.5 & 12800 & 65.60 & 51.81 \\
Kanade 25Hz & 25 & 12800 & 61.53 & 51.15 \\
\bottomrule
\end{tabular}
\end{table}

\subsection{Metric correlation analysis}
\label{appendix:correlation}

\textbf{High phonetic discriminability is not a necessary condition for high lexical availability.} Although Kanade models do not get the best phonetic metrics (as seen in Table~\ref{table:phonetic-metrics}), they still achieve the SOTA performance on downstream ASR (as seen in Table~\ref{table:asr}). This observation leads us to further investigate the correlation between different linguistic metrics. 

\begin{figure}[t]
\centering
\includegraphics[width=\linewidth]{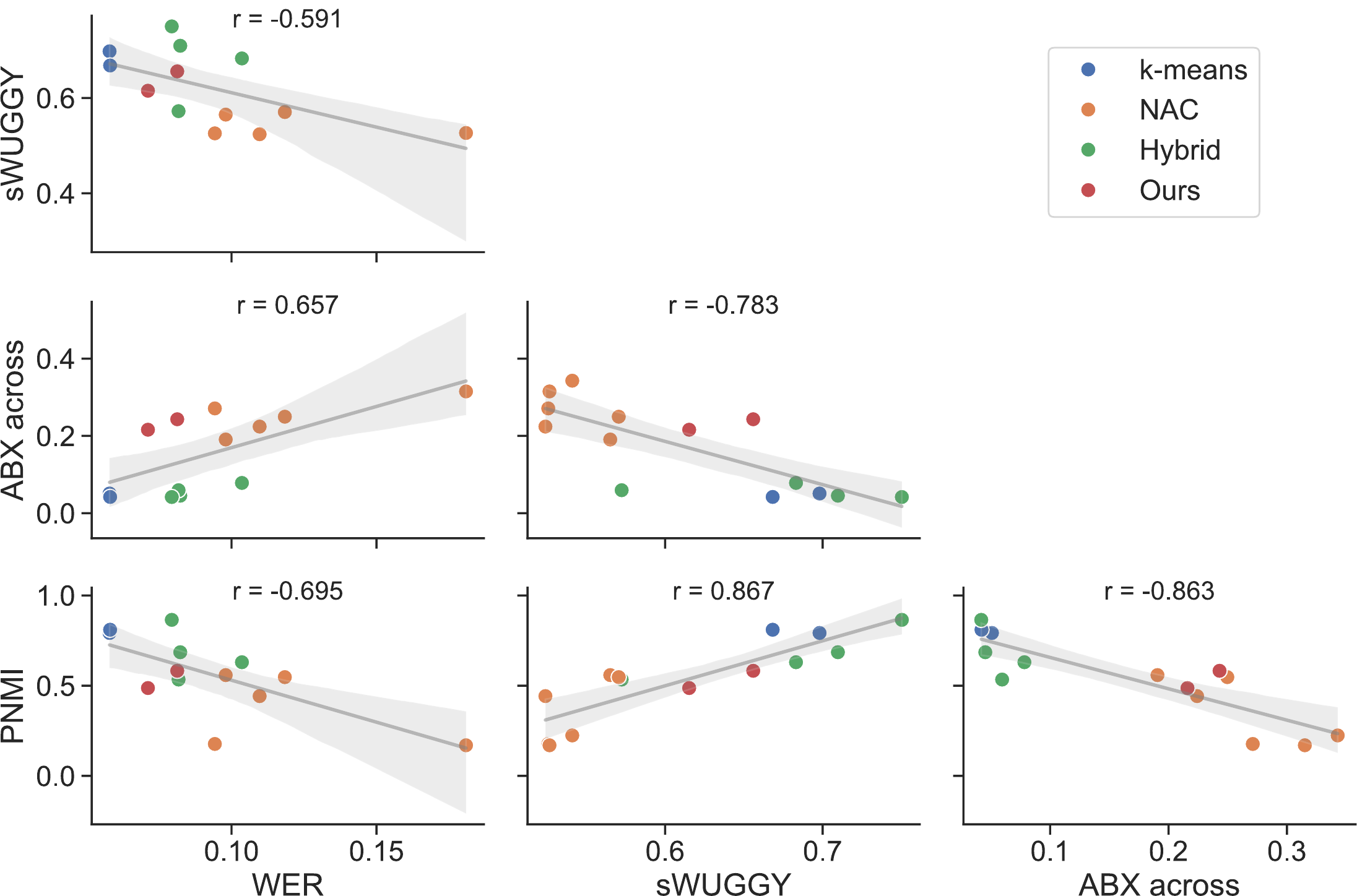}
\caption{\textbf{Correlation among metrics of lexical and phonetic performance.} Lexical metrics include downstream ASR WER (Table~\ref{table:asr}) and sWUGGY in spoken language modeling (Table~\ref{table:slm-preliminary}). Phonetic metrics include ABX across and PNMI (Table~\ref{table:phonetic-metrics}). Coarse model groupings are included for readability.}
\label{fig:correlation}
\end{figure}

The results are shown in Figure~\ref{fig:correlation}, where we observe correlation between downstream WER and phonetic metrics (also reported by \citet{chang2024dcspin}).

~\\

However, the relationship is not perfect. Notably, in the ABX-WER plot (second row, first column):

\begin{itemize}[leftmargin=1em, nolistsep, topsep=0pt]
    \item Our models (red dots) are significantly higher than the regression line, which means they are better at providing lexical information than the models with similar phonetic performance.
    \item Hybrid codec models (green dots) are significantly lower than the regression line, which means they fail to achieve word error rates typical of models with similar ABX scores (k-means).
\end{itemize}

The ABX-sWUGGY plot (second row, second column) also shows our models achieve noticeably better sWUGGY scores than NACs, despite having similar ABX scores. \citet{huang2024repcodec} also report that the relationship between phonetic discriminability and downstream performance is not strict: they recorded these scores during training and observed that PNMI scores peaked early then decreased in parallel with downstream WER.

These results suggest that ABX and PNMI, originally designed for acoustic unit discovery, may not be sufficient to measure token quality for downstream modeling. Kanade tokens perform similarly to NAC tokens on these metrics, but perform similarly to k-means tokens on lexical tasks. We hypothesize that Kanade tokens may contain more non-phonetic linguistic information that can help identify words or might have a less well-behaved continuous embedding space. However, without further investigation, we cannot make a decisive conclusion.

\subsection{Out-of-distribution reconstruction}
\label{appendix:ood}

\begin{table*}[p]
\caption{\textbf{OOD reconstruction results}. Evaluation on various out-of-distribution (OOD) datasets. \dag\ indicates models trained on relevant data (e.g., noisy data or Japanese). Includes only the best models from the reconstruction results. For all results see Tables~\ref{table:ood-full-1} and~\ref{table:ood-full-2}.}
\label{table:ood}
\scriptsize
\centering
\begin{tabular}{
    m{8em} 
    S[table-format=2.1,round-mode=places,round-precision=1,detect-weight]
    S[table-format=2.1,round-mode=places,round-precision=1,detect-weight]
    S[table-format=1.2,round-mode=places,round-precision=2,detect-weight]
    S[table-format=1.2,round-mode=places,round-precision=2,detect-weight]
    S[table-format=1.2,round-mode=places,round-precision=2,detect-weight]
    S[table-format=1.2,round-mode=places,round-precision=2,detect-weight]
}
\toprule
\multirow{2}{*}{\textbf{Model}} & \multicolumn{2}{c}{\textbf{Intelligibility}} & \multicolumn{1}{c}{\textbf{Quality}} & \multicolumn{1}{c}{\textbf{Speaker}} & \multicolumn{2}{c}{\textbf{Prosody}} \\
\cmidrule(lr){2-3} \cmidrule(lr){4-4} \cmidrule(lr){5-5} \cmidrule(lr){6-7}
 & {WER$\downarrow$} & {CER$\downarrow$} & {UTMOS$\uparrow$} & {SIM$\uparrow$} & {F0Corr$\uparrow$} & {F0RMSE$\downarrow$} \\
\midrule
\multicolumn{7}{c}{Gigaspeech~\citep{chen2021gigaspeech} \textit{(noisy speech)}} \\
\noalign{\vskip -1pt}
\cmidrule(lr){1-7}
Ground Truth & 9.7 & 5.1 & 2.84 & \NA & \NA & \NA \\
X-Codec 2$^\dagger$ & 11.5 & 6.3 & 2.99 & 0.97 & 0.87 & 0.08 \\
BiCodec$^\dagger$ & 11.9 & 6.6 & 3.07 & 0.96 & 0.87 & 0.08 \\
PAST \textcolor[gray]{0.6}{1:8} & 10.9 & 6.0 & 3.09 & 0.98 & 0.89 & 0.07 \\
DualCodec \textcolor[gray]{0.6}{1:8}$^\dagger$ & 11.0 & 6.0 & 3.11 & 0.98 & 0.84 & 0.08 \\
WavTokenizer & 33.9 & 21.9 & 2.64 & 0.88 & 0.82 & 0.10 \\
StableCodec$^\dagger$ & 27.1 & 16.3 & 3.51 & 0.90 & 0.84 & 0.09 \\
Kanade 12.5Hz & 16.2 & 9.3 & 3.25 & 0.95 & 0.74 & 0.13 \\
Kanade 25Hz & 11.3 & 6.2 & 3.27 & 0.96 & 0.81 & 0.09 \\
\cmidrule(lr){1-7}
\multicolumn{7}{c}{Salmon Sentiment Consistency~\citep{maimon2025salmon} \textit{(emotional)}} \\
\noalign{\vskip -1pt}
\cmidrule(lr){1-7}
Ground Truth & 2.9 & 1.0 & 3.79 & \NA & \NA & \NA \\
\; w/ change & 4.9 & 1.6 & 3.62 & \NA & \NA & \NA \\
X-Codec 2$^\dagger$ & 3.8 & 1.2 & 3.77 & 0.97 & 0.85 & 0.09 \\
\; w/ change & 5.7 & 2.2 & 3.67 & 0.97 & 0.89 & 0.11 \\
BiCodec$^\dagger$ & 5.4 & 1.7 & 3.84 & 0.98 & 0.81 & 0.10 \\
\; w/ change & 6.0 & 2.6 & 3.73 & 0.97 & 0.90 & 0.11 \\
PAST \textcolor[gray]{0.6}{1:8} & 3.0 & 1.0 & 3.91 & 0.99 & 0.85 & 0.09 \\
\; w/ change & 4.2 & 1.7 & 3.77 & 0.98 & 0.90 & 0.08 \\
DualCodec \textcolor[gray]{0.6}{1:8}$^\dagger$ & 3.6 & 1.1 & 3.91 & 0.98 & 0.88 & 0.08 \\
\; w/ change & 4.4 & 1.8 & 3.76 & 0.98 & 0.90 & 0.10 \\
WavTokenizer & 14.5 & 7.7 & 3.21 & 0.90 & 0.74 & 0.12 \\
\; w/ change & 17.5 & 9.7 & 3.13 & 0.90 & 0.82 & 0.16 \\
StableCodec$^\dagger$ & 14.8 & 7.2 & 4.08 & 0.93 & 0.81 & 0.12 \\
\; w/ change & 18.0 & 9.3 & 4.03 & 0.92 & 0.84 & 0.12 \\
Kanade 12.5Hz & 6.4 & 2.3 & 3.83 & 0.95 & 0.66 & 0.19 \\
\; w/ change & 7.0 & 3.1 & 3.83 & 0.94 & 0.67 & 0.22 \\
Kanade 25Hz & 4.4 & 1.5 & 3.85 & 0.96 & 0.73 & 0.16 \\
\; w/ change & 4.7 & 1.9 & 3.88 & 0.96 & 0.75 & 0.18 \\
\cmidrule(lr){1-7}
\multicolumn{7}{c}{Japanese Versatile Speech~\citep{takamichi2019jvs} \textit{(unseen language speech)}} \\
\noalign{\vskip -1pt}
\cmidrule(lr){1-7}
Ground Truth & 4.6 & 2.5 & 3.63 & \NA & \NA & \NA \\
X-Codec 2$^\dagger$ & 5.4 & 2.9 & 3.59 & 0.98 & 0.89 & 0.10 \\
BiCodec & 5.7 & 3.1 & 3.73 & 0.98 & 0.86 & 0.10 \\
PAST \textcolor[gray]{0.6}{1:8} & 5.2 & 2.8 & 3.62 & 0.98 & 0.88 & 0.09 \\
DualCodec \textcolor[gray]{0.6}{1:8}$^\dagger$ & 5.0 & 2.8 & 3.67 & 0.99 & 0.81 & 0.09 \\
WavTokenizer & 18.2 & 11.3 & 2.92 & 0.88 & 0.82 & 0.14 \\
StableCodec & 25.0 & 16.5 & 3.83 & 0.91 & 0.90 & 0.10 \\
Kanade 12.5Hz & 12.2 & 7.2 & 3.77 & 0.94 & 0.70 & 0.21 \\
Kanade 25Hz & 5.6 & 3.0 & 3.72 & 0.97 & 0.84 & 0.17 \\
\cmidrule(lr){1-7}
\multicolumn{7}{c}{English Read by Japanese~\citep{nakagawa2007erj} \textit{(accented speech)}} \\
\noalign{\vskip -1pt}
\cmidrule(lr){1-7}
Ground Truth & 14.9 & 8.0 & 3.73 & \NA & \NA & \NA \\
X-Codec 2$^\dagger$ & 20.7 & 11.3 & 3.69 & 0.97 & 0.86 & 0.08 \\
BiCodec & 21.4 & 11.7 & 3.76 & 0.97 & 0.86 & 0.07 \\
PAST \textcolor[gray]{0.6}{1:8} & 25.3 & 14.1 & 3.65 & 0.97 & 0.85 & 0.07 \\
DualCodec \textcolor[gray]{0.6}{1:8}$^\dagger$ & 17.1 & 9.4 & 3.71 & 0.98 & 0.86 & 0.07 \\
WavTokenizer & 51.7 & 31.6 & 3.06 & 0.91 & 0.82 & 0.08 \\
StableCodec & 51.4 & 29.3 & 4.03 & 0.91 & 0.87 & 0.06 \\
Kanade 12.5Hz & 33.8 & 18.6 & 3.78 & 0.95 & 0.80 & 0.09 \\
Kanade 25Hz & 22.9 & 12.3 & 3.75 & 0.96 & 0.86 & 0.07 \\
\bottomrule
\end{tabular}
\end{table*}

\begin{figure*}[p]
\centering
\includegraphics[width=0.48\linewidth]{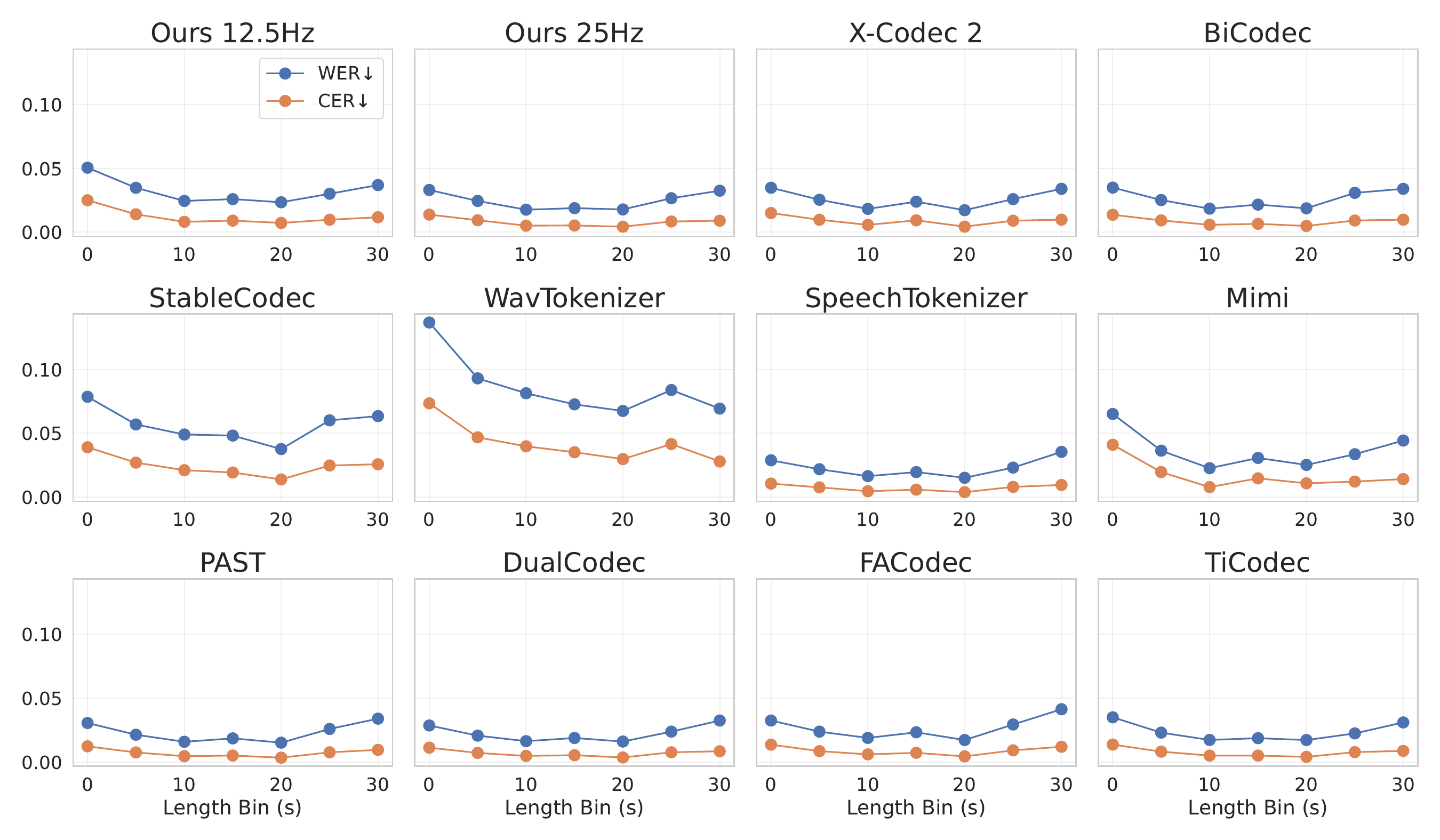}
\includegraphics[width=0.63\linewidth]{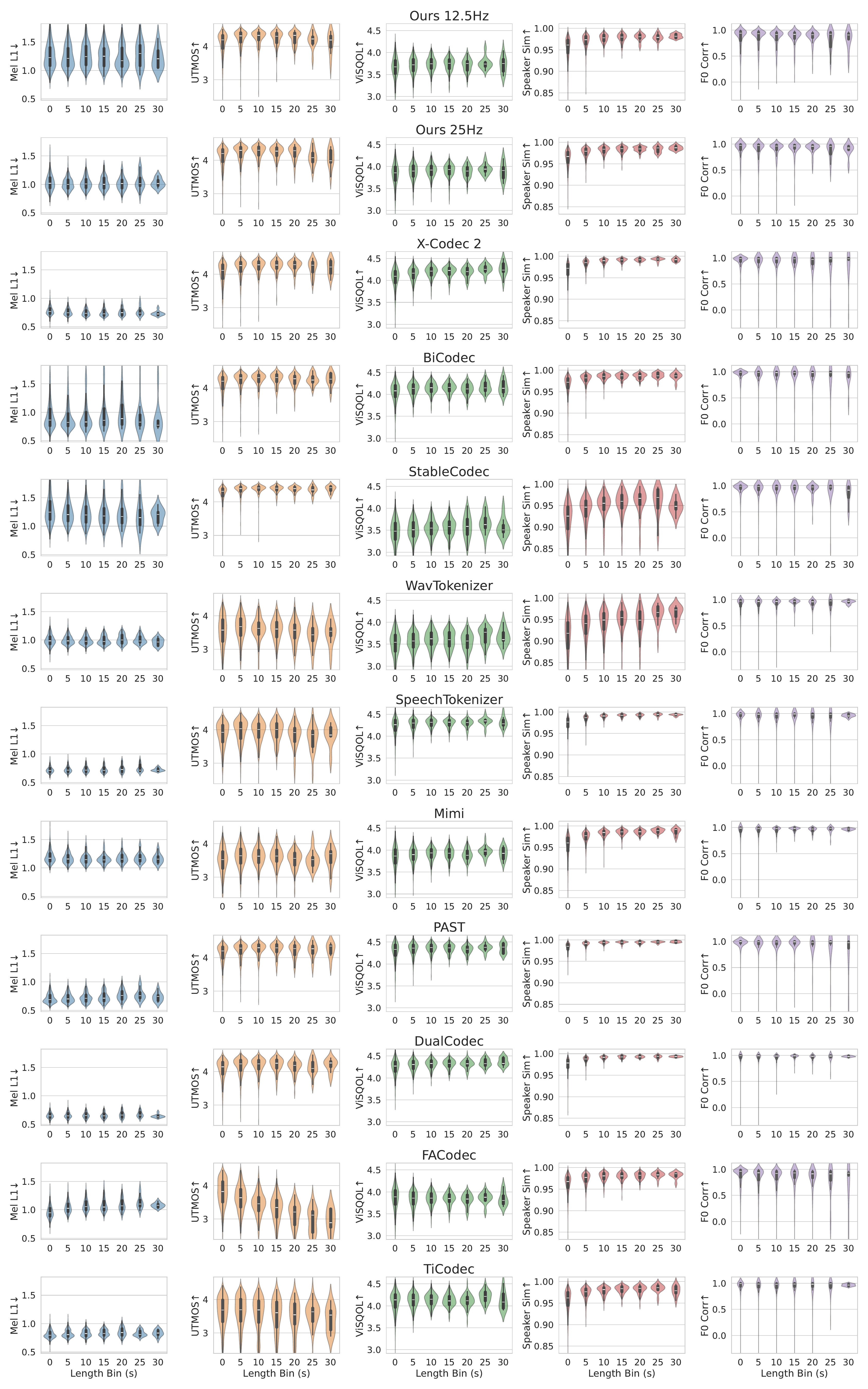}
\caption{\textbf{Reconstruction metrics on different audio length bins}}
\label{fig:length-gen}
\end{figure*}

We reconstructed randomly sampled utterances from out-of-distribution datasets. Objective metrics are shown in Table~\ref{table:ood}. We included the best baselines from Table~\ref{table:reconstruction}. 

Kanade performs competitively in various scenarios despite being trained on very little data. Still, subjectively, we found phone substitution errors to be common in the poorest reconstructions; phone deletion also occurred with some frequency.

\textbf{Noisy/Spontaneous}\; We tested noisy speech by sampling utterances with at least two words from GigaSpeech~\citep{chen2021gigaspeech}. Transcripts were preprocessed to remove punctuation and other tags before computing WER. Listening to the reconstructions, we found that background music and noise was partially captured by the global embedding, as expected. Even though Kanade has only seen read English speech, it maintains some of the best WERs in this condition.

\textbf{Emotional}\; We tested emotional speech using the sentiment consistency subset of Salmon~\citep{maimon2025salmon}, which is derived from the emotional speech dataset Expresso~\citep{nguyen2023expresso}. Whispered samples were excluded. Each track has a consistent version (only one speech style/emotion) and an inconsistent version (speech style/emotion changes within the utterance). This dataset was chosen to test how Kanade encodes large changes in speaking style across and within utterances, which is not seen in the read English audio it was trained on.
We report results for each version separately (\textit{w/ change} indicates results for the inconsistent track). Subjectively, reconstructions of consistent samples were good. Inconsistent samples had some leakage of style into the global embedding, causing them to become more uniform upon resynthesis. Nearly all metrics are degraded in the inconsistent case. Interestingly, even speech tokenizers without disentanglement also suffered under this condition.

\textbf{Unseen Language}\; We also tested on Japanese, which was not seen during training. Transcripts and ASR results were normalized to phonological script before comparison. While the 25Hz variant is quite good (22\% relative increase in WER, in line with results on English speech), the 12.5Hz variant performs poorly (165\% relative increase in WER). Subjectively, it sounds slightly accented.

\textbf{Accented}\; Finally, we reconstructed Japanese-accented English speech using sentence samples from ERJ~\citep{nakagawa2007erj}. Since segmentals in this dataset are not always clearly in an English phonetic category, we suspect that our discretization step may incorrectly categorize them and eliminate the ambiguity that would normally allow an ASR model to recover using its language modeling capabilities. No speech tokenizer did well on these utterances.

These experiments show that Kanade performs competitively in various scenarios despite being trained on very little data. The consistency experiment shows that large changes in vocal quality do not have a large detrimental effect on intelligibility.

These reconstructions can be found on our demo page\footnote{\url{https://frothywater.github.io/kanade-tokenizer/}}.

\subsection{Length generalization}
\label{appendix:length-gen}

A good speech tokenizer should work on audio that is longer than the sequences it was trained on. We test the length generalization performance of several high-performing baselines and Kanade on LibriSpeech \texttt{test-clean} using reconstruction metrics, binned by audio length. The bin width is 5 seconds, with the final bin including all samples more than 30 seconds long. The results are shown in Figure~\ref{fig:length-gen}. Most models perform well. Kanade models (trained on 5.76s segments) show consistent performance on every metric even at 6x the audio length, indicating excellent length generalization.

\subsection{Chunked encoding and decoding}
\label{appendix:chunked-encoding-decoding}

\subsubsection{Chunked resynthesis}
To further show that it is possible to encode audio of arbitrary length with Kanade, we report metrics for a chunked resynthesis experiment. We randomly select 100 samples (25-35 minutes per sample) from Libri-Light~\citep{kahn2020librilight}. Each is encoded into 5.76-second segments that overlap for 1.44 seconds. We decode with either (1) simple mean over all global embeddings of the chunks, or (2) an exponential moving average ($\alpha = 0.8$). We then combine the tracks using a 10ms crossfade.

To evaluate, we select 1000 randomly-selected LibriHeavy~\citep{kang2024libriheavy} segments that occur within the 100 LibriLight samples. The corresponding segments are cut out from the resynthesized audio and used to compute the metrics in Table~\ref{table:chunked-resynthesis}.

This experiment shows that a single global embedding (the simple mean) is enough to encode large amounts of audio with high fidelity.

\begin{table}[t]
\caption{\textbf{Chunked Resynthesis Results}. Includes resynthesis using the average global embedding for all chunks or an exponential moving average (EMA).}
\label{table:chunked-resynthesis}
\scriptsize
\centering
\begin{tabular}{
    m{6.5em} 
    S[table-format=2.1,round-mode=places,round-precision=1,detect-weight]
    S[table-format=2.1,round-mode=places,round-precision=1,detect-weight]
    S[table-format=1.2,round-mode=places,round-precision=2,detect-weight]
    S[table-format=1.2,round-mode=places,round-precision=2,detect-weight]
    S[table-format=1.2,round-mode=places,round-precision=2,detect-weight]
    S[table-format=1.2,round-mode=places,round-precision=2,detect-weight]
    S[table-format=1.2,round-mode=places,round-precision=2,detect-weight]
}
\toprule
\multirow{2}{*}{\textbf{Model}} & \multicolumn{2}{c}{\textbf{Intelligibility}} & \multicolumn{1}{c}{\textbf{Quality}} & \multicolumn{2}{c}{\textbf{Speaker}} & \multicolumn{2}{c}{\textbf{Prosody}} \\
\cmidrule(lr){2-3} \cmidrule(lr){4-4} \cmidrule(lr){5-6} \cmidrule(lr){7-8}
& {WER$\downarrow$} & {CER$\downarrow$} & {UTMOS$\uparrow$} & {SIM$\uparrow$} & {F0Corr$\uparrow$} & {F0RMSE$\downarrow$} \\
\midrule
Ground Truth & 4.02 & 1.72 & 3.887 & \NA & \NA & \NA \\
\midrule
Kanade 12.5Hz \\
\quad Simple mean & 5.67 & 2.62 & 3.982 & 0.974 & 0.753 & 0.127 \\
\quad EMA & 5.46 & 2.48 & 3.969 & 0.979 & 0.835 & 0.092 \\
Kanade 25Hz \\
\quad Simple mean & 4.70 & 2.15 & 4.003 & 0.979 & 0.782 & 0.123 \\
\quad EMA & 4.56 & 2.08 & 3.988 & 0.983 & 0.854 & 0.080 \\
\bottomrule
\end{tabular}
\end{table}

\subsubsection{Chunked streaming for SLMs}
For an interactive speech language model, it is not necessary to compute any global embeddings as they are not used as input and the output speech is synthesized using a constant global embedding.

\textbf{Rough latency estimate}\; As Table~\ref{appendix:efficiency} shows, Kanade is extremely fast. Therefore, input latency is dominated by the amount of padding necessary on the end of the audio input to get a reasonable representation. According to work by \citet{meng2025effective}, we can estimate that we need a 400ms lookahead and 2 seconds of history to get SSL features that are reasonably accurate (and in turn, good tokens). For synthesis of the SLM output, it is not clear how much lookahead is necessary, but we conservatively estimate that it is the same as the input, 400ms. Therefore, the minimum theoretical latency is 800ms plus Kanade encoding time (2.4s times Kanade's encoding-decoding RTF of 0.0011 is 3ms) and SLM latency. A streaming variant would decrease the necessary lookahead and decrease latency substantially.

\subsection{Codebook utilization}

\begin{table}[t]
\caption{\textbf{Normalized entropy of different speech tokenizers.} For multi-layer codecs, only the linguistically dense token layer (usually the first RVQ layer) is used.}
\label{table:codebook-entropy}
\scriptsize
\centering
\begin{tabular}{lc}
\toprule
\textbf{Tokenizer} & \textbf{Entropy} \\
\midrule
BiCodec & 0.995 \\
ST & 0.984 \\
X-Codec 2 & 0.965 \\
StableCodec & 0.962 \\
FACodec & 0.953 \\
DualCodec & 0.923 \\
Mimi & 0.914 \\
WavTokenizer & 0.885 \\
PAST & 0.824 \\
TiCodec & 0.623 \\
\midrule
Kanade 12.5Hz & 0.976 \\
Kanade 25Hz & 0.974 \\
\bottomrule
\end{tabular}
\end{table}

To test the codebook utilization of baselines and Kanade, we calculate normalized entropy as:
$$\text{Normalized Entropy}=-\frac{1}{\log N} \sum_{x=1}^{N} p(x)\log(p(x)),$$
where $N$ is the codebook size and $p(x)$ denotes the probability distribution of extracted codes at codebook index $x$. Values are between 0 and 1. Higher values indicates better codebook utilization. Note that for codebook-free models such as StableCodec, X-Codec 2 and Kanade, the codebook here refers to the effective codebook produced by FSQ indices. We estimate this using the tokens extracted from each tokenizer on LibriSpeech \texttt{test-clean}. The results are shown in Table~\ref{table:codebook-entropy}. Most of the tested model have good codebook utilization. The normalized entropy values of Kanade content tokens are over 97\%, indicating excellent coding efficiency.

\section{Experiment details}
\subsection{Model and training}
\label{appendix:model}
\textbf{Model Details}\; The content encoder and feature decoder are 6-layer, 12-head, 768-dim LLaMA~\citep{grattafiori2024llama3herdmodels}-style transformers with rotary position embeddings (RoPE)~\citep{su2024roformer}, 2048-dim SwiGLU~\citep{shazeer2020glu} feed-forward networks, and local attention (window size 125). 

The FSQ~\citep{mentzer2024finite}\footnote{FSQ typically works in a very low-dimensional space, and partitions it using a simple fixed grid. To perform FSQ on a vector, we first project it into that lower-dimensional space. Then for each dimension we 1) squash it using a scaled $\tanh$ such it lies in a bounded range $(a, b)$ of the reals; and 2) round it to the nearest integer. There are a finite number of integers between $a$ and $b$ and these correspond to our quantization levels. Since the scaling factor of the squashing function can be chosen, we can freely choose the number of levels for each dimension.} module uses 5 dimensions with levels of $[8, 8, 8, 5, 5]$, equivalent to a codebook of 12,800 tokens. This results in bitrates of 171bps and 341bps for the 12.5Hz and 25Hz models, respectively. 

The global branch uses a 4-layer, 384-dim ConvNeXt~\citep{liu2022convnet} encoder with attentive statistics pooling~\citep{okabe2018attentive}\footnote{Attentive statistics pooling passes the input with $d$ features to a simple convolutional network to weight each element of the input. The mean and standard deviation are then computed over the time dimension for each feature, producing one vector of dimension $2d$ for the entire sequence. The result is passed through a linear layer to obtain the final dimension of the global embedding and then layer normalized.} to produce a 128-dim global embedding. 

The token module is a 6-layer, 12-head, 768-dim transformer (window size 31/65 for 12.5/25Hz model), and the mel module is a 6-layer, 8-head, 512-dim transformer (window size 65) with adaLN-Zero~\citep{peebles2023scalable} conditioning. 
The post-net consists of 5 convolutional layers with a kernel size of 7 and 256 channels.
The Mel spectrograms use 100 bins, 1024-point FFT, and 256 hop length, consistent with Vocos~\citep{siuzdak2024vocos}. 

The discriminator used in post-training is a multi-band spectrogram discriminator directly applied on our generated mel spectrogram, adapted from DAC~\citep{kumar2023high}. It splits the mel bins into 5 bands, and processed each band using 5 convolution layers with kernel size of $[3,3]$ and 64 channels. For a higher-level overview, see Section~\ref{section:gan-losses}.

The resulting 12.5Hz model has 120M training parameters and 207M total parameters (containing 73M from WavLM Base+ and 13.5M from Vocos). The 25Hz variant has 118M training parameters and 205M total parameters.

\textbf{Training Details}\; We train the models for 150k steps with a batch size of 128 using randomly chunked 5.76-second audio segments. The SSL feature and mel-spectrogram reconstruction losses are weighted equally ($\alpha=1$).
We optimize with AdamW~\citep{loshchilov2018decoupled} ($\beta_1=0.9, \beta_2=0.99$, weight decay 1e-4) and a cosine learning rate schedule with a peak of 2e-4 and a 10\% warmup. 

In the GAN post-training phase, the weights for adversarial loss and feature matching loss are $\beta = 1/30$ and $\gamma = 1/3$, respectively.
We use a constant learning rate of 4e-5 and select the final checkpoint based on validation mel L1 loss and subjective quality.

All models are trained with \texttt{bfloat16} mixed precision and FlashAttention 2~\citep{dao2023flashattention} for efficiency.
Training takes approximately 32 hours on one NVIDIA 5090 GPU in total.

\subsection{Training data and inference efficiency}
\label{appendix:efficiency}

\begin{table*}[tb]
\caption{\textbf{Training data and inference efficiency.} Real time factor (RTF) indicates ratios of processing time for either encoding (En) or decoding (De) to input audio length, measured on a NVIDIA A6000. Relative efficiency is calculated on full passes.}
\label{table:efficiency}
\scriptsize
\centering
\setlength{\tabcolsep}{4pt}
\begin{tabular}{lcccccccc}
\toprule
\textbf{Model} & \textbf{Params} & \textbf{Dataset} & \makecell{\textbf{Sample} \\ \textbf{Rate}} & \makecell{\textbf{Data Size} \\ (hours)} & \makecell{\textbf{RTF} \\ (En) $\downarrow$} & \makecell{\textbf{RTF} \\ (De) $\downarrow$} & \makecell{\textbf{Relative} \\ \textbf{Efficiency} $\uparrow$}\\
\midrule
StableCodec & 953M & Libri-Light + MLS & 16 kHz & 105k & 0.0028 & 0.0028 & 0.18x \\
X-Codec 2 & 823M & Emilia + MLS & 16 kHz & 150k & 0.0181 & 0.0016 & 0.05x \\
PAST & 184M & LibriSpeech + TIMIT & 16 kHz & 1.0k & 0.0012 & 0.0007 & 0.53x \\
BiCodec & 156M & LibriSpeech + Emilia & 16 kHz & 3k & 0.0045 & 0.0030 & 0.13x \\
ST & 104M & LibriSpeech & 16 kHz & 1k & 0.0010 & 0.0008 & 0.59x \\
FACodec & 102M & Libri-Light & 16 kHz & 60k & 0.0035 & 0.0067 & 0.10x \\
DualCodec & 84M & Emilia & 24 kHz & 100k & 0.0078 & 0.0011 & 0.11x \\
WavTokenizer & 81M & LibriTTS & 24 kHz & 0.6k & 0.0003 & 0.0003 & 1.67x \\
Mimi & 79M & \NA & 24 kHz & \NA & 0.0007 & 0.0006 & 0.77x \\
TiCodec & 63M & LibriTTS & 24 kHz & 0.6k & 0.0021 & 0.0028 & 0.21x \\
\midrule
Kanade 12.5Hz & 207M & LibriTTS & 24 kHz & 0.6k & 0.0009 & 0.0002 & 1.00x \\
Kanade 25Hz & 205M & LibriTTS & 24 kHz & 0.6k & 0.0009 & 0.0002 & 1.00x \\
\bottomrule
\end{tabular}
\end{table*}

One benefit of using SSL features to train a speech tokenizer is data efficiency. As shown in Table~\ref{table:efficiency}, we use much less data than comparable models (0.6k vs. X-Codec 2's 150k hours). Kanade models are relatively lightweight, with one fifth the parameters of StableCodec, but the 25Hz version still obtains a competitive MUSHRA subjective quality score (75.0 vs. StableCodec's 79.3). Inference speed is also excellent, surpassing all baselines except WavTokenizer.

\subsection{Downstream model configurations}
\label{appendix:downstream}

Our transformer-based downstream models all share similar backbones to the ones used in our main model. They are 12-layer, 12-head, 768-dim LLaMA-style transformers with 85M parameters (excluding embedding and output projection layers). Downstream transformers are configured as decoder-only with causal attention. We use AdamW~\citep{loshchilov2018decoupled} ($\beta_1=0.9, \beta_2=0.999$, weight decay 1e-3) and a cosine learning rate schedule with a peak of 2e-4 and a 10\% warmup. 

\textbf{ASR}\; Before training, we train SentencePiece~\citep{kudo2018sentencepiece} text tokenizers on LibriSpeech or SwitchBoard\footnote{\url{https://huggingface.co/datasets/hhoangphuoc/switchboard}} transcripts with a vocabulary size of 5,000. The transformer model is trained for 100k steps, with each batch of tokens extracted from 240 seconds of speech. The training sequences are in the format \texttt{<speech><BOS><text><EOS>} (as illustrated in Figure~\ref{fig:downstream-models}). Cross-entropy loss is calculated on text tokens only. For RVQ models, we use multiple embedding layers, concatenate the resulting embeddings along the feature dimension, then project them back to the original dimension via a linear layer. For the continuous reference model, we use the average of layer 6 and 9 features as input. We use label smoothing of 0.1. After training, we select the best checkpoint with the lowest validation loss to test the final WER. During testing, we set beam size as 8, length penalty to 1.0, and patience factor to 2.0.

\textbf{TTS}\; Before training, we run grapheme-to-phoneme on LibriTTS transcripts to get all phonemes using SoundChoice~\citep{ploujnikov22soundchoice}. A transformer is trained on the LibriTTS training sets for 200k steps, each step with a batch of tokens extracted from 120 seconds of speech. The sequence format is \texttt{<speaker embedding><phonemes><BOS><speech><EOS>} (as illustrated in Figure~\ref{fig:downstream-models}), where the cross-entropy loss during training is calculated on speech tokens only. For RVQ models, we combine all code indices of the used RVQ codebooks to create the token vocabulary following AudioLM~\citep{borsos2023audiolm}. For example, if a tokenizer uses two 1024-code codebooks, then the first codebook has indices $[0, 1023]$, and the second has indices $[1024, 2047]$, forming a final vocabulary of size 2048. We use the last checkpoint for evaluation. During inference, we set the temperature to 1.0 and use top-p sampling with $p = 0.9$. We omit FACodec as a baseline here as its token rate is too high (480Hz).

\textbf{Speaker discriminators}\; For speaker tasks, we use ECAPA-TDNN~\citep{desplanques2020ecapa} backbones. Following RawNet3~\citep{jung2022pushing}, the token embedding dimension, hidden dimension, and final embedding dimension are 192, 1024, and 192, respectively. We train for 50k steps on batches of 64 randomly cropped 3-second samples. The scale and margin in AAM-Softmax loss~\citep{deng2019arcface} are set to 30 and 0.3, respectively. We use AdamW ($\beta_1=0.9, \beta_2=0.999$, weight decay 1e-3) with a constant learning rate 3e-4. We select the best checkpoint with the lowest validation loss for evaluation.

For RVQ models, the token embeddings from different token layers are concatenated and projected back to the original dimension via a linear layer. Global embedding is projected to the token embedding dimension and added to the token embeddings at each time step. For TiCodec and BiCodec, which produce fixed-length global tokens, we allocate individual embedding layers for each token index and aggregate those embeddings. 
To evaluate only on global embedding, we replace the ECAPA-TDNN with a 3-layer MLP. The hidden dimension is 768.

\subsection{Subjective listening test}
\label{appendix:subjective}
We conduct Multiple Stimuli with Hidden Reference and Anchor (MUSHRA) subjective listening tests using webMUSHRA~\citep{schoeffler2018webmushra}.

For the reconstruction quality scores, we ask the subjects to judge \textit{``unnatural or robotic-sounding speech; muffled or distorted sound; the rhythm and melody of the voice sounding unnatural; the speaker's voice sounding different; and incorrect words or slurred pronunciation''}. The ground truth is shown as reference. 10 audio samples of between 3–6 seconds are randomly selected from LibriSpeech \texttt{test-clean}.

For TTS, we prepare for two different tests. (1) In the speech quality test, we ask the subjects to judge \textit{``robotic artifacts, static noise, muffled sound; slurred pronunciation or unclear speech''} and
ignore \textit{``the speaker's emotion, rhythm, speed, pitch, or intonation''}. (2) In the prosody naturalness test, we ask the subjects to judge \textit{``the melody of the voice (intonation), correct stress on words, natural speed, and logical pauses (rhythm)''} and ignore \textit{``audio quality issues such as static, robotic buzzing, or muffled sounds''}. There is no reference shown in the TTS tests. 10 audio samples are randomly selected from the LibriTTS \texttt{test-clean}.

For the VC speaker similarity test, we ask the subjects to judge \textit{``if the sample sounds exactly like the same person as the reference''}. The reference speech from the target speaker is shown as reference. 10 audio samples are randomly selected from the VCTK subset.

For all tests, the ground truth is included as a hidden condition. Each sample is scored by at least 25 people. Since it is difficult for participants to score many models at once, we divide the models into groups with roughly balanced quality composition based on objective metrics. We removed outlier participants from the collected data and calibrated the groups by the mean reference scores among the groups.
Lowpass-filtered anchors are not used.

We use bootstrapping~\citep{mendoncca2018statistical} with 1000 iterations to estimate the median scores for each models and report 95\% confidence intervals (see Section~\ref{section:full-results}).

\subsection{Baselines}
\label{appendix:baselines}

\textbf{SpeechTokenizer}~\citep{zhang2024speechtokenizer}\; A hybrid speech codec that distills HuBERT~\citep{hsu2021hubert} features into the first of 8 RVQ layers. By doing this, SpeechTokenizer makes its first layer more like HuBERT features, making them a suitable alternative to SSL k-means tokens for spoken language modeling. The rest of the token layers encode the rest of the information necessary for reconstruction. It is one of the earliest hybrid speech codecs. The token rate per layer is 50Hz and the codebook size is 1024. We use the \texttt{hubert\_avg} checkpoint. SpeechTokenizer and other RVQ-based models introduced below support variable bitrates by using only the first $N$ token layers, thanks to random quantizer dropout training.\footnote{\url{https://github.com/ZhangXInFD/SpeechTokenizer}}

\textbf{PAST}~\citep{har2025past}\; A hybrid speech codec that distills phoneme labels and text into the first quantization layer, similar to SpeechTokenizer. The token rate per layer is 50Hz and the codebook size is 1024. We use the non-streamable checkpoint.\footnote{\url{https://github.com/slp-rl/PAST}}

\textbf{Mimi}~\citep{defossez2024moshi}\; A streaming hybrid speech codec that distills WavLM~\citep{chen2022wavlm} features into the first quantization layer, similar to SpeechTokenizer. The difference is Mimi uses a separate VQ layer for distillation alongside 7 normal RVQ layers. The token rate per layer is 12.5Hz and the codebook size is 2048.\footnote{\url{https://huggingface.co/kyutai/mimi}}

\textbf{DualCodec}~\citep{li25dualcodec}\; A hybrid speech codec that incorporates SSL features by compressing w2v-BERT 2.0~\citep{barrault2023seamless} features with a ConvNeXt~\citep{liu2022convnet}-based VQ-VAE and using the quantized latents as RVQ 1. A separate encoder is applied to the waveform to produce an acoustic embedding. RVQ 1 is decoded and subtracted from the acoustic embedding. The remaining 7 RVQ layers quantize the residual. The token rate per layer is 12.5Hz and the codebook size is 16,384 for the first layer and 4,096 for the rest.\footnote{\url{https://github.com/jiaqili3/dualcodec}}

\textbf{TiCodec}~\citep{ren2024fewer}\; A disentangled speech codec that separates time-varying content and time-invariant information. During training, it extracts global information from a randomly sampled reference segment different than the input one, then uses Group Vector Quantization (GVQ) to produces 8 global tokens (codebook size 1024). To enforce disentanglement, it applies Siamese-like consistency loss between the global representations from the original segment and the reference segment. It has three separately trained variants for 1, 2, and 4 RVQ layers in the content branch. The token rate per layer is 75Hz and the codebook size is 1024. We mostly use the variant with 4 layers, except the places where single layer is preferred (e.g., PNMI metric, SLM experiments).\footnote{\url{https://github.com/y-ren16/TiCodec}}

\textbf{FACodec}~\citep{ju2024naturalspeech}\; A disentangled speech codec that separates prosody, phonetic content, and speaker identity. It produces 6 RVQ layers: 1 for prosody (supervised by F0), 2 for phonetic content (supervised by phonemes sequences) and 3 for residual details. It also produces a global speaker embedding learned by speaker supervision. To enforce disentanglement, FACodec applies gradient reversal layers on each branch. The token rate per layer is 80Hz and the codebook size is 1024. The speaker embedding is 256-dim.\footnote{\url{https://github.com/open-mmlab/Amphion/tree/main/models/codec/ns3_codec}}

\textbf{BiCodec}~\citep{wang2025spark}\; A single-layer disentangled speech codec that separates linguistic content and global information. It uses wav2vec 2.0~\citep{baevski2020wav2vec} features as main input and extracts global tokens from mel spectrogram to represent constant acoustics such as speaker timbre. It uses cross attention mechanism similar to Q-former on ECAPA-TDNN features to extract 32 global tokens, which are then quantized by FSQ (codebook size 4096). The decoder reconstructs both the waveform and SSL features. The token rate is 50Hz and the codebook size is 8192.\footnote{\url{https://github.com/SparkAudio/Spark-TTS}}

\textbf{WavTokenizer}~\citep{ji2024wavtokenizer}\; A single-layer neural audio codec uses several techniques to improve codebook utilization, such as k-means initialization and dead code random restart. It also uses a ConvNeXt~\citep{liu2022convnet} backbone and predicts Short-Time Fourier Transform magnitude and phase values instead of waveform. The token rate is 40Hz and the codebook size is 4096. We use the speech-only checkpoint \texttt{small-600-24k-4096}.\footnote{\url{https://github.com/jishengpeng/WavTokenizer}}

\textbf{StableCodec}~\citep{parker2024scaling}\; A large transformer-based single-layer neural audio codec that uses a novel post-hoc residual formulation of FSQ~\citep{mentzer2024finite}. They show transformers' great scalability in speech coding and reach very low a bitrate of 400bps. It represents one of the earliest speech codecs with a transformer-based architecture. In their official repository, the authors further fine-tune the model using CTC loss on phonemes to enhance lexical information. Following their recommendation, we use this fine-tuned checkpoint \texttt{stable-codec-speech-16k}. The token rate is 25Hz and the codebook size is 46656.\footnote{\url{https://github.com/Stability-AI/stable-codec}}

\textbf{X-Codec 2}~\citep{ye2025llasa}\; A single-layer neural audio codec that adds a parallel VQ-VAE for w2v-BERT 2.0~\citep{barrault2023seamless} feature reconstruction alongside the original acoustic VQ-VAE. Frozen SSL and acoustic features are projected and concatenated into a shared space that is quantized using FSQ~\citep{mentzer2024finite}. The token rate is 50Hz and the codebook size is 65536.\footnote{\url{https://huggingface.co/HKUSTAudio/xcodec2}}

We don't include the earlier codecs such as EnCodec~\citep{defossez2023high} and DAC~\citep{kumar2023high} because (1) they mainly focus on high-quality general audio coding, while we focus on speech-only tokenizers that have potential for speech language modeling; (2) they need more tokens to reconstruct good quality audio, with the lowest bitrates starting from 1.5kbps, which is impractical for speech LMs; and (3) their approaches are already well represented and improved on in later works such as SpeechTokenizer, PAST, Mimi, and DualCodec.

\clearpage
\onecolumn
\subsection{Full results}
\label{section:full-results}
\begin{table}[h]
\caption{\textbf{Full Speech reconstruction results.} Grouped by model family. Bold numbers indicate the best performance in that column.}
\label{table:reconstruction-all}
\scriptsize
\centering
\begin{tabular}{
    m{7em} 
    >{\centering\arraybackslash}m{3em} 
    >{\centering\arraybackslash}m{3em} 
    S[table-format=2.1,detect-weight]
    S[table-format=2.1,detect-weight]
    S[table-format=2.2,detect-weight]
    S[table-format=1.2,detect-weight]
    S[table-format=1.2,detect-weight]
    S[table-format=1.2,detect-weight]
    S[table-format=1.2,detect-weight]
}
\toprule
\multirow{2}{*}{\textbf{Model}} & \multirow{2}{*}{\textbf{Bitrate}} & \multirow{2}{*}{\makecell{\textbf{Token} \\ \textbf{Rate}}} & \multicolumn{2}{c}{\textbf{Intelligibility}} & \multicolumn{2}{c}{\textbf{Quality}} & \multicolumn{1}{c}{\textbf{Speaker}} & \multicolumn{2}{c}{\textbf{Prosody}} \\
\cmidrule(lr){4-5} \cmidrule(lr){6-7} \cmidrule(lr){8-8} \cmidrule(lr){9-10}
 &  &  & {WER$\downarrow$} & {CER$\downarrow$} & {MUSHRA$\uparrow$} & {UTMOS$\uparrow$} & {SIM$\uparrow$} & {F0Corr$\uparrow$} & {F0RMSE$\downarrow$} \\
\midrule
\rowcolor{gray!10} Ground Truth & {--} & -- & 1.9 & 0.6 & 78.0 & 4.07 & {--} & {--} & {--} \\
\rowcolor{gray!10} Cont. 50Hz & {--} & -- & 2.0 & 0.6 & 76.7 & 3.90 & 0.99 & 0.94 & 0.04 \\
\rowcolor{gray!10} KM 25Hz & 341 & 25 & 2.7 & 1.0 & 72.4 & 4.07 & 0.96 & 0.67 & 0.15 \\
\rowcolor{gray!10} KM 12.5Hz & 171 & 12.5 & 3.0 & 1.1 & 72.1 & 4.04 & 0.96 & 0.66 & 0.15 \\
\midrule
FACodec* \textcolor[gray]{0.6}{1:6} & 4800 & 480 & \bfseries 2.1 & \bfseries 0.7 & 81.4 & 4.11 & 0.98 & 0.94 & \bfseries 0.04 \\
FACodec* \textcolor[gray]{0.6}{1:3} & 2400 & 240 & 2.4 & 0.8 & {--} & 3.62 & 0.97 & 0.85 & 0.08 \\
PAST \textcolor[gray]{0.6}{1:8} & 4000 & 400 & \bfseries 2.1 & \bfseries 0.7 & \bfseries 82.4 & 4.18 & \bfseries 0.99 & 0.92 & \bfseries 0.04 \\
PAST \textcolor[gray]{0.6}{1:4} & 2000 & 200 & 2.4 & 0.9 & {--} & 3.88 & 0.98 & 0.89 & 0.06 \\
PAST \textcolor[gray]{0.6}{1:2} & 1000 & 100 & 3.1 & 1.2 & {--} & 2.45 & 0.88 & 0.39 & 0.31 \\
ST \textcolor[gray]{0.6}{1:8} & 4000 & 400 & \bfseries 2.1 & \bfseries 0.7 & 76.0 & 3.90 & 0.98 & 0.92 & 0.05 \\
ST \textcolor[gray]{0.6}{1:4} & 2000 & 200 & 2.6 & 0.9 & 74.2 & 3.56 & 0.96 & 0.88 & 0.07 \\
ST \textcolor[gray]{0.6}{1:2} & 1000 & 100 & 3.6 & 1.4 & {--} & 2.28 & 0.90 & 0.78 & 0.11 \\
TiCodec* \textcolor[gray]{0.6}{1:4} & 3000 & 300 & 2.3 & 0.8 & {--} & 3.60 & 0.97 & 0.91 & 0.05 \\
TiCodec* \textcolor[gray]{0.6}{1:2} & 1500 & 150 & 3.7 & 1.6 & {--} & 3.43 & 0.94 & 0.88 & 0.07 \\
TiCodec* \textcolor[gray]{0.6}{1:1} & 750 & 75 & 9.3 & 4.8 & {--} & 3.17 & 0.91 & 0.85 & 0.08 \\
Mimi \textcolor[gray]{0.6}{1:8} & 1100 & 100 & 3.7 & 1.9 & {--} & 3.56 & 0.97 & 0.93 & 0.05 \\
Mimi \textcolor[gray]{0.6}{1:4} & 550 & 50 & 7.7 & 5.1 & {--} & 3.02 & 0.93 & 0.87 & 0.09 \\
Mimi \textcolor[gray]{0.6}{1:2} & 275 & 25 & 14.7 & 10.8 & {--} & 2.39 & 0.86 & 0.60 & 0.17 \\
DualCodec \textcolor[gray]{0.6}{1:8} & 1225 & 100 & \bfseries 2.1 & \bfseries 0.7 & 75.6 & 4.12 & 0.98 & \bfseries 0.95 & \bfseries 0.04 \\
DualCodec \textcolor[gray]{0.6}{1:4} & 625 & 50 & 2.6 & 0.9 & {--} & 4.07 & 0.97 & 0.93 & 0.05 \\
DualCodec \textcolor[gray]{0.6}{1:2} & 325 & 25 & 3.7 & 1.5 & 72.4 & 3.67 & 0.94 & 0.91 & 0.07 \\
X-Codec 2 & 800 & 50 & 2.5 & 0.9 & 77.0 & 4.13 & 0.98 & 0.90 & 0.06 \\
BiCodec* & 650 & 50 & 2.5 & 0.9 & 75.0 & 4.18 & 0.98 & 0.91 & 0.05 \\
WavTokenizer & 480 & 40 & 9.4 & 4.7 & 72.1 & 3.57 & 0.92 & 0.91 & 0.07 \\
StableCodec & 388 & 25 & 5.7 & 2.6 & 79.3 & \bfseries 4.31 & 0.93 & 0.91 & 0.05 \\
\midrule
Kanade* 25Hz & 341 & 25 & 2.4 & 0.8 & 75.0 & 4.16 & 0.97 & 0.88 & 0.07 \\
Kanade* 12.5Hz & 171 & 12.5 & 3.3 & 1.3 & 74.6 & 4.17 & 0.97 & 0.85 & 0.10 \\
\bottomrule
\end{tabular}
{
\begin{flushleft}
Models marked with * also use a fixed-size representation for reconstruction. FACodec: 8192 bits (256-dim \texttt{fp32}), TiCodec: 80 bits (8 tokens), BiCodec: 384 bits (32 tokens), and Kanade: 4096 bits (128-dim \texttt{fp32}).
\end{flushleft}
}
\end{table}
\begin{table}[H]
\caption{\textbf{Full voice conversion results}}
\label{table:vc-full}
\scriptsize
\centering
\begin{tabular}{
    m{6.5em}
    S[table-format=2.1,round-mode=places,round-precision=1,detect-weight]
    S[table-format=2.1,round-mode=places,round-precision=1,detect-weight]
    S[table-format=1.2,round-mode=places,round-precision=2,detect-weight]
    S[table-format=2.1,round-mode=places,round-precision=1,detect-weight]
    S[table-format=1.2,round-mode=places,round-precision=2,detect-weight]
}
\toprule
\multirow{2}{*}{\textbf{Model}} & \multicolumn{2}{c}{\textbf{Intelligibility}} & \textbf{Quality} & \textbf{Speaker} & \textbf{Prosody} \\
\cmidrule(lr){2-3} \cmidrule(lr){4-4} \cmidrule(lr){5-5} \cmidrule(lr){6-6}
& {WER$\downarrow$} & {CER$\downarrow$} & {UTMOS$\uparrow$} & {EER$\uparrow$} & {F0Corr$\uparrow$} \\
\midrule
\rowcolor{gray!10} Ground Truth & 0.0 & 0.0 & 4.08 & \NA & \NA \\
\rowcolor{gray!10} LinearVC & 0.6 & 0.2 & 3.94 & 29.7 & 0.62 \\
\rowcolor{gray!10} FreeVC & 0.6 & 0.3 & 3.99 & 29.0 & 0.67 \\
\rowcolor{gray!10} CosyVoice 2 & 1.1 & 0.5 & 4.11 & 31.0 & 0.64 \\
\midrule
PAST \textcolor[gray]{0.6}{1:8} & 22.9 & 15.1 & 1.84 & 8.2 & 0.20 \\
PAST \textcolor[gray]{0.6}{1:4} & 13.3 & 8.3 & 1.80 & 5.4 & 0.17 \\
PAST \textcolor[gray]{0.6}{1:2} & 6.6 & 3.8 & 1.69 & 3.9 & 0.17 \\
ST \textcolor[gray]{0.6}{1:8} & 74.7 & 61.7 & 1.54 & 10.6 & 0.19 \\
ST \textcolor[gray]{0.6}{1:4} & 35.2 & 26.1 & 1.62 & 8.9 & 0.19 \\
ST \textcolor[gray]{0.6}{1:2} & 10.6 & 6.0 & 1.52 & 5.8 & 0.22 \\
TiCodec \textcolor[gray]{0.6}{1:4} & \bfseries 0.5 & \bfseries 0.2 & 3.32 & 5.4 & 0.77 \\
TiCodec \textcolor[gray]{0.6}{1:2} & 3.4 & 1.9 & 3.13 & 5.7 & 0.74 \\
TiCodec \textcolor[gray]{0.6}{1:1} & 10.2 & 6.1 & 3.25 & 8.9 & 0.64 \\
Mimi \textcolor[gray]{0.6}{1:8} & 120.3 & 86.8 & 3.09 & \bfseries 38.5 & 0.24 \\
Mimi \textcolor[gray]{0.6}{1:4} & 110.8 & 84.6 & 2.15 & 15.2 & 0.21 \\
Mimi \textcolor[gray]{0.6}{1:2} & 102.4 & 85.3 & 1.59 & 5.1 & 0.18 \\
DualCodec \textcolor[gray]{0.6}{1:8} & 21.5 & 12.9 & 2.50 & 6.8 & 0.54 \\
DualCodec \textcolor[gray]{0.6}{1:4} & 8.5 & 4.6 & 2.88 & 7.1 & 0.56 \\
DualCodec \textcolor[gray]{0.6}{1:2} & 4.4 & 2.3 & 3.07 & 5.8 & 0.62 \\
BiCodec & 1.2 & 0.6 & 3.84 & 18.5 & 0.61 \\
FACodec \textcolor[gray]{0.6}{1:6} & 0.7 & 0.4 & 3.765 & 15.2 & \bfseries 0.789 \\
FACodec \textcolor[gray]{0.6}{1:3} & 0.8 & 0.4 & 3.45 & 18.6 & 0.66 \\
\midrule
Kanade 25Hz & 0.7 & 0.3 & 4.16 & 30.7 & 0.71 \\
Kanade 12.5Hz & 1.6 & 0.7 & \bfseries 4.17 & 32.0 & 0.64 \\
\bottomrule
\end{tabular}
\end{table}

\begin{table}[h]
\begin{minipage}{0.45\textwidth}
\caption{\textbf{Full reconstruction MUSHRA results} with 95\% confidence intervals.}
\label{table:reconstruction-mushra}
\scriptsize
\centering
\begin{tabular}{
    m{8em}
    S[table-format=2.1,round-mode=places,round-precision=1,detect-weight]
    S[table-format=2.1,round-mode=places,round-precision=1,detect-weight]
    S[table-format=2.1,round-mode=places,round-precision=1,detect-weight]
}
\toprule
\textbf{Model} & \textbf{$-$} & \textbf{Median} & \textbf{$+$} \\
\midrule
Ground Truth & 76.0 & 78.0 & 80.0 \\
Cont. 50Hz & 72.1 & 76.7 & 80.3 \\
KM 12.5Hz & 66.9 & 72.1 & 76.2 \\
KM 25Hz & 68.2 & 72.4 & 76.1 \\
\midrule
ST \textcolor[gray]{0.6}{1:8} & 72.0 & 76.0 & 78.0 \\
ST \textcolor[gray]{0.6}{1:4} & 64.9 & 74.2 & 78.8 \\
DualCodec \textcolor[gray]{0.6}{1:8} & 73.5 & 75.6 & 80.9 \\
DualCodec \textcolor[gray]{0.6}{1:2} & 68.2 & 72.4 & 75.6 \\
FACodec & 77.8 & 81.4 & 83.4 \\
PAST & 78.3 & 82.4 & 84.5 \\
StableCodec & 75.2 & 79.3 & 81.4 \\
X-Codec 2 & 74.0 & 77.0 & 80.0 \\
BiCodec & 72.0 & 75.0 & 79.0 \\
WavTokenizer & 65.9 & 72.1 & 76.2 \\
\midrule
Kanade 12.5Hz & 70.3 & 74.5 & 77.7 \\
w/o GAN & 59.0 & 69.0 & 74.0 \\
\; w/o Dual-branch & 15.0 & 24.0 & 46.5 \\
\; w/o SSL Recon. & 57.5 & 68.5 & 75.5 \\
\; w/o End-to-End & 51.1 & 60.7 & 66.6 \\
\; w/o FSQ & 31.4 & 43.7 & 55.9 \\
Kanade 25Hz & 72.0 & 75.0 & 78.0 \\
\; w/o GAN & 66.0 & 70.3 & 75.6 \\
\bottomrule
\end{tabular}
\end{minipage}\hfill
\begin{minipage}{0.45\textwidth}
\caption{\textbf{Full voice conversion speaker similarity MUSHRA results} with 95\% confidence intervals.}
\label{table:vc-mushra}
\scriptsize
\centering
\begin{tabular}{
    m{8em}
    S[table-format=2.1,round-mode=places,round-precision=1,detect-weight]
    S[table-format=2.1,round-mode=places,round-precision=1,detect-weight]
    S[table-format=2.1,round-mode=places,round-precision=1,detect-weight]
}
\toprule
\textbf{Model} & \textbf{$-$} & \textbf{Median} & \textbf{$+$} \\
\midrule
Ground Truth & 72.0 & 74.5 & 77.0 \\
LinearVC & 69.3 & 73.4 & 78.1 \\
FreeVC & 71.0 & 74.5 & 77.5 \\
CosyVoice 2 & 73.0 & 76.0 & 79.0 \\
\midrule
ST & 25.0 & 35.0 & 47.5 \\
Mimi & 77.6 & 81.7 & 85.9 \\
DualCodec & 34.0 & 52.0 & 68.0 \\
FACodec & 51.7 & 62.6 & 69.3 \\
PAST & 15.5 & 23.3 & 50.7 \\
TiCodec & 57.0 & 68.0 & 73.0 \\
BiCodec & 66.7 & 71.4 & 75.5 \\
\midrule
Kanade 12.5Hz & 72.4 & 77.6 & 81.7 \\
Kanade 25Hz & 73.4 & 77.1 & 80.7 \\
\bottomrule
\end{tabular}
\end{minipage}
\end{table}

\begin{table}[h]
\begin{minipage}{0.45\textwidth}
\caption{\textbf{Full TTS speech quality MUSHRA results} with 95\% confidence intervals.}
\label{table:tts-quality-mushra}
\scriptsize
\centering
\begin{tabular}{
    m{8em}
    S[table-format=2.1,round-mode=places,round-precision=1,detect-weight]
    S[table-format=2.1,round-mode=places,round-precision=1,detect-weight]
    S[table-format=2.1,round-mode=places,round-precision=1,detect-weight]
}
\toprule
\textbf{Model} & \textbf{$-$} & \textbf{Median} & \textbf{$+$} \\
\midrule
Ground Truth & 72.0 & 74.9 & 77.1 \\
KM 25Hz & 71.5 & 74.9 & 79.3 \\
KM 12.5Hz & 67.0 & 72.0 & 78.5 \\
CosyVoice 2 & 74.9 & 77.1 & 79.3 \\
\midrule
ST & 69.0 & 75.0 & 78.0 \\
Mimi & 71.5 & 74.9 & 78.2 \\
DualCodec & 69.0 & 73.0 & 78.0 \\
PAST & 70.4 & 74.9 & 79.3 \\
TiCodec & 71.5 & 73.8 & 77.1 \\
StableCodec & 64.0 & 71.0 & 77.0 \\
X-Codec 2 & 68.0 & 72.0 & 78.0 \\
BiCodec & 69.8 & 73.8 & 76.0 \\
WavTokenizer & 68.0 & 74.5 & 79.0 \\
\midrule
Kanade 12.5Hz & 72.6 & 77.1 & 79.3 \\
Kanade 25Hz & 67.0 & 73.0 & 80.0 \\
\bottomrule
\end{tabular}
\end{minipage}\hfill
\begin{minipage}{0.45\textwidth}
\caption{\textbf{Full TTS prosody naturalness MUSHRA results} with 95\% confidence intervals.}
\label{table:tts-prosody-mushra}
\scriptsize
\centering
\begin{tabular}{
    m{8em}
    S[table-format=2.1,round-mode=places,round-precision=1,detect-weight]
    S[table-format=2.1,round-mode=places,round-precision=1,detect-weight]
    S[table-format=2.1,round-mode=places,round-precision=1,detect-weight]
}
\toprule
\textbf{Model} & \textbf{$-$} & \textbf{Median} & \textbf{$+$} \\
\midrule
Ground Truth & 78.9 & 80.9 & 83.0 \\
KM 12.5Hz & 60.0 & 67.0 & 73.0 \\
KM 25Hz & 69.8 & 75.9 & 78.9 \\
CosyVoice 2 & 80.9 & 83.0 & 85.5 \\
\midrule
ST & 75.0 & 79.0 & 81.0 \\
Mimi & 66.8 & 73.9 & 78.4 \\
DualCodec & 74.0 & 80.0 & 83.0 \\
PAST & 72.9 & 78.4 & 81.5 \\
TiCodec & 65.8 & 72.9 & 78.9 \\
StableCodec & 58.0 & 66.0 & 74.5 \\
X-Codec 2 & 75.0 & 78.0 & 81.0 \\
BiCodec & 73.9 & 78.9 & 82.0 \\
WavTokenizer & 73.0 & 77.0 & 80.0 \\
\midrule
Kanade 12.5Hz & 73.9 & 77.9 & 80.9 \\
Kanade 25Hz & 78.0 & 81.0 & 83.0 \\
\bottomrule
\end{tabular}
\end{minipage}
\end{table}

\begin{table}[tb]
\caption{\textbf{Full OOD reconstruction results (Part I).} Evaluation on noisy (Gigaspeech) and emotional (Salmon) speech. \dag\ indicates models trained on noisy data.}
\label{table:ood-full-1}
\tiny
\centering
\begin{tabular}{
    m{8em} 
    S[table-format=2.1,round-mode=places,round-precision=1,detect-weight]
    S[table-format=2.1,round-mode=places,round-precision=1,detect-weight]
    S[table-format=1.2,round-mode=places,round-precision=2,detect-weight]
    S[table-format=1.2,round-mode=places,round-precision=2,detect-weight]
    S[table-format=1.2,round-mode=places,round-precision=2,detect-weight]
    S[table-format=1.2,round-mode=places,round-precision=2,detect-weight]
}
\toprule
\multirow{2}{*}{\textbf{Model}} & \multicolumn{2}{c}{\textbf{Intelligibility}} & \multicolumn{1}{c}{\textbf{Quality}} & \multicolumn{1}{c}{\textbf{Speaker}} & \multicolumn{2}{c}{\textbf{Prosody}} \\
\cmidrule(lr){2-3} \cmidrule(lr){4-4} \cmidrule(lr){5-5} \cmidrule(lr){6-7}
 & {WER$\downarrow$} & {CER$\downarrow$} & {UTMOS$\uparrow$} & {SIM$\uparrow$} & {F0Corr$\uparrow$} & {F0RMSE$\downarrow$} \\
\midrule
\multicolumn{7}{c}{Gigaspeech~\citep{chen2021gigaspeech} \textit{(noisy speech)}} \\
\noalign{\vskip -1pt}
\cmidrule(lr){1-7}
Ground Truth & 9.7 & 5.1 & 2.84 & \NA & \NA & \NA \\
FACodec \textcolor[gray]{0.6}{1:6} & 11.3 & 6.3 & 2.85 & 0.97 & 0.88 & 0.07 \\
PAST \textcolor[gray]{0.6}{1:8} & 10.9 & 6.0 & 3.09 & 0.98 & 0.89 & 0.07 \\
PAST \textcolor[gray]{0.6}{1:4} & 12.6 & 7.1 & 2.70 & 0.96 & 0.81 & 0.11 \\
PAST \textcolor[gray]{0.6}{1:2} & 18.5 & 11.1 & 1.78 & 0.85 & 0.27 & 0.34 \\
ST \textcolor[gray]{0.6}{1:8} & 11.8 & 6.6 & 2.60 & 0.97 & 0.88 & 0.08 \\
ST \textcolor[gray]{0.6}{1:4} & 14.7 & 8.8 & 2.41 & 0.93 & 0.83 & 0.10 \\
ST \textcolor[gray]{0.6}{1:2} & 21.4 & 13.1 & 1.71 & 0.85 & 0.75 & 0.13 \\
TiCodec \textcolor[gray]{0.6}{1:4} & 12.4 & 7.0 & 2.45 & 0.95 & 0.86 & 0.08 \\
TiCodec \textcolor[gray]{0.6}{1:2} & 18.5 & 11.5 & 2.35 & 0.91 & 0.83 & 0.09 \\
TiCodec \textcolor[gray]{0.6}{1:1} & 31.4 & 21.0 & 2.25 & 0.88 & 0.74 & 0.13 \\
Mimi \textcolor[gray]{0.6}{1:8}$^\dagger$ & 12.3 & 7.0 & 2.71 & 0.96 & 0.85 & 0.09 \\
Mimi \textcolor[gray]{0.6}{1:4}$^\dagger$ & 16.0 & 9.6 & 2.37 & 0.93 & 0.79 & 0.11 \\
Mimi \textcolor[gray]{0.6}{1:2}$^\dagger$ & 22.6 & 14.2 & 1.98 & 0.85 & 0.58 & 0.17 \\
DualCodec \textcolor[gray]{0.6}{1:8}$^\dagger$ & 11.0 & 6.0 & 3.11 & 0.98 & 0.84 & 0.08 \\
DualCodec \textcolor[gray]{0.6}{1:4}$^\dagger$ & 12.3 & 7.0 & 3.07 & 0.96 & 0.83 & 0.09 \\
DualCodec \textcolor[gray]{0.6}{1:2}$^\dagger$ & 15.8 & 9.3 & 2.78 & 0.93 & 0.81 & 0.10 \\
X-Codec 2$^\dagger$ & 11.5 & 6.3 & 2.99 & 0.97 & 0.87 & 0.08 \\
BiCodec$^\dagger$ & 11.9 & 6.6 & 3.07 & 0.96 & 0.87 & 0.08 \\
WavTokenizer & 33.9 & 21.9 & 2.64 & 0.88 & 0.82 & 0.10 \\
StableCodec$^\dagger$ & 27.1 & 16.3 & 3.51 & 0.90 & 0.84 & 0.09 \\
Kanade 12.5Hz & 16.2 & 9.3 & 3.25 & 0.95 & 0.74 & 0.13 \\
Kanade 25Hz & 11.3 & 6.2 & 3.27 & 0.96 & 0.81 & 0.09 \\
\cmidrule(lr){1-7}
\multicolumn{7}{c}{Salmon Sentiment~\citep{maimon2025salmon} \textit{(emotional)}} \\
\noalign{\vskip -1pt}
\cmidrule(lr){1-7}
Ground Truth & 2.9 & 1.0 & 3.79 & \NA & \NA & \NA \\
\; w/ change & 4.9 & 1.6 & 3.62 & \NA & \NA & \NA \\
FACodec \textcolor[gray]{0.6}{1:6} & 3.8 & 1.2 & 3.87 & 0.98 & 0.92 & 0.08 \\
\; w/ change & 4.4 & 1.8 & 3.77 & 0.98 & 0.90 & 0.09 \\
FACodec \textcolor[gray]{0.6}{1:3} & 3.9 & 1.4 & 3.32 & 0.97 & 0.79 & 0.15 \\
\; w/ change & 5.9 & 2.2 & 3.34 & 0.96 & 0.78 & 0.17 \\
PAST \textcolor[gray]{0.6}{1:8} & 3.0 & 1.0 & 3.91 & 0.99 & 0.85 & 0.09 \\
\; w/ change & 4.2 & 1.7 & 3.77 & 0.98 & 0.90 & 0.08 \\
PAST \textcolor[gray]{0.6}{1:4} & 3.9 & 1.4 & 3.46 & 0.96 & 0.86 & 0.10 \\
\; w/ change & 5.4 & 2.0 & 3.32 & 0.95 & 0.85 & 0.13 \\
PAST \textcolor[gray]{0.6}{1:2} & 6.9 & 3.0 & 1.96 & 0.72 & 0.20 & 0.49 \\
\; w/ change & 6.1 & 2.8 & 1.92 & 0.68 & 0.10 & 0.49 \\
ST \textcolor[gray]{0.6}{1:8} & 3.9 & 1.2 & 3.53 & 0.97 & 0.86 & 0.10 \\
\; w/ change & 4.3 & 1.7 & 3.42 & 0.97 & 0.86 & 0.10 \\
ST \textcolor[gray]{0.6}{1:4} & 5.2 & 1.7 & 3.15 & 0.92 & 0.79 & 0.11 \\
\; w/ change & 7.4 & 3.7 & 3.11 & 0.91 & 0.88 & 0.12 \\
ST \textcolor[gray]{0.6}{1:2} & 9.4 & 3.8 & 2.30 & 0.82 & 0.72 & 0.16 \\
\; w/ change & 10.6 & 5.2 & 2.30 & 0.82 & 0.80 & 0.15 \\
TiCodec \textcolor[gray]{0.6}{1:4} & 3.9 & 1.3 & 3.44 & 0.96 & 0.92 & 0.10 \\
\; w/ change & 4.5 & 1.9 & 3.28 & 0.96 & 0.86 & 0.11 \\
TiCodec \textcolor[gray]{0.6}{1:2} & 6.0 & 2.7 & 3.07 & 0.92 & 0.88 & 0.10 \\
\; w/ change & 8.0 & 4.3 & 3.00 & 0.91 & 0.81 & 0.12 \\
TiCodec \textcolor[gray]{0.6}{1:1} & 16.7 & 9.3 & 2.98 & 0.88 & 0.75 & 0.16 \\
\; w/ change & 19.0 & 10.6 & 2.83 & 0.87 & 0.78 & 0.15 \\
Mimi \textcolor[gray]{0.6}{1:8}$^\dagger$ & 4.1 & 1.8 & 3.22 & 0.96 & 0.82 & 0.11 \\
\; w/ change & 5.9 & 3.0 & 3.09 & 0.95 & 0.79 & 0.14 \\
Mimi \textcolor[gray]{0.6}{1:4}$^\dagger$ & 6.1 & 2.9 & 2.75 & 0.91 & 0.75 & 0.14 \\
\; w/ change & 8.5 & 4.6 & 2.65 & 0.91 & 0.77 & 0.16 \\
Mimi \textcolor[gray]{0.6}{1:2}$^\dagger$ & 13.2 & 8.4 & 2.18 & 0.83 & 0.48 & 0.23 \\
\; w/ change & 14.7 & 9.1 & 2.18 & 0.82 & 0.53 & 0.24 \\
DualCodec \textcolor[gray]{0.6}{1:8}$^\dagger$ & 3.6 & 1.1 & 3.91 & 0.98 & 0.88 & 0.08 \\
\; w/ change & 4.4 & 1.8 & 3.76 & 0.98 & 0.90 & 0.10 \\
DualCodec \textcolor[gray]{0.6}{1:4}$^\dagger$ & 4.7 & 1.8 & 3.88 & 0.97 & 0.78 & 0.12 \\
\; w/ change & 5.3 & 2.4 & 3.77 & 0.97 & 0.91 & 0.10 \\
DualCodec \textcolor[gray]{0.6}{1:2}$^\dagger$ & 6.8 & 2.9 & 3.46 & 0.94 & 0.80 & 0.13 \\
\; w/ change & 7.0 & 3.4 & 3.41 & 0.94 & 0.81 & 0.15 \\
X-Codec 2$^\dagger$ & 3.8 & 1.2 & 3.77 & 0.97 & 0.85 & 0.09 \\
\; w/ change & 5.7 & 2.2 & 3.67 & 0.97 & 0.89 & 0.11 \\
BiCodec$^\dagger$ & 5.4 & 1.7 & 3.84 & 0.98 & 0.81 & 0.10 \\
\; w/ change & 6.0 & 2.6 & 3.73 & 0.97 & 0.90 & 0.11 \\
WavTokenizer & 14.5 & 7.7 & 3.21 & 0.90 & 0.74 & 0.12 \\
\; w/ change & 17.5 & 9.7 & 3.13 & 0.90 & 0.82 & 0.16 \\
StableCodec$^\dagger$ & 14.8 & 7.2 & 4.08 & 0.93 & 0.81 & 0.12 \\
\; w/ change & 18.0 & 9.3 & 4.03 & 0.92 & 0.84 & 0.12 \\
Kanade 12.5Hz & 6.4 & 2.3 & 3.83 & 0.95 & 0.66 & 0.19 \\
\; w/ change & 7.0 & 3.1 & 3.83 & 0.94 & 0.67 & 0.22 \\
Kanade 25Hz & 4.4 & 1.5 & 3.85 & 0.96 & 0.73 & 0.16 \\
\; w/ change & 4.7 & 1.9 & 3.88 & 0.96 & 0.75 & 0.18 \\
\bottomrule
\end{tabular}
\end{table}
\begin{table}[tb]
\caption{\textbf{Full OOD reconstruction results (Part II).} Evaluation on unseen language (JVS) and accented speech (ERJ). \dag\ indicates models trained on Japanese.}
\label{table:ood-full-2}
\tiny
\centering
\begin{tabular}{
    m{8em} 
    S[table-format=2.1,round-mode=places,round-precision=1,detect-weight]
    S[table-format=2.1,round-mode=places,round-precision=1,detect-weight]
    S[table-format=1.2,round-mode=places,round-precision=2,detect-weight]
    S[table-format=1.2,round-mode=places,round-precision=2,detect-weight]
    S[table-format=1.2,round-mode=places,round-precision=2,detect-weight]
    S[table-format=1.2,round-mode=places,round-precision=2,detect-weight]
}
\toprule
\multirow{2}{*}{\textbf{Model}} & \multicolumn{2}{c}{\textbf{Intelligibility}} & \multicolumn{1}{c}{\textbf{Quality}} & \multicolumn{1}{c}{\textbf{Speaker}} & \multicolumn{2}{c}{\textbf{Prosody}} \\
\cmidrule(lr){2-3} \cmidrule(lr){4-4} \cmidrule(lr){5-5} \cmidrule(lr){6-7}
& {WER$\downarrow$} & {CER$\downarrow$} & {UTMOS$\uparrow$} & {SIM$\uparrow$} & {F0Corr$\uparrow$} & {F0RMSE$\downarrow$} \\
\midrule
\multicolumn{7}{c}{JVS~\citep{takamichi2019jvs} \textit{(unseen language)}} \\
\noalign{\vskip -1pt}
\cmidrule(lr){1-7}
Ground Truth & 4.6 & 2.5 & 3.63 & \NA & \NA & \NA \\
FACodec \textcolor[gray]{0.6}{1:6} & 5.1 & 2.8 & 3.69 & 0.97 & 0.90 & 0.09 \\
FACodec \textcolor[gray]{0.6}{1:3} & 6.4 & 3.5 & 2.89 & 0.95 & 0.79 & 0.18 \\
PAST \textcolor[gray]{0.6}{1:8} & 5.2 & 2.8 & 3.62 & 0.98 & 0.88 & 0.09 \\
PAST \textcolor[gray]{0.6}{1:4} & 7.3 & 4.1 & 2.73 & 0.91 & 0.80 & 0.16 \\
PAST \textcolor[gray]{0.6}{1:2} & 17.0 & 10.8 & 1.63 & 0.64 & 0.16 & 0.53 \\
ST \textcolor[gray]{0.6}{1:8} & 5.7 & 3.2 & 3.32 & 0.96 & 0.86 & 0.10 \\
ST \textcolor[gray]{0.6}{1:4} & 7.8 & 4.6 & 2.87 & 0.90 & 0.82 & 0.12 \\
ST \textcolor[gray]{0.6}{1:2} & 16.0 & 10.4 & 2.02 & 0.80 & 0.81 & 0.15 \\
TiCodec \textcolor[gray]{0.6}{1:4} & 5.6 & 3.1 & 3.21 & 0.95 & 0.86 & 0.10 \\
TiCodec \textcolor[gray]{0.6}{1:2} & 8.5 & 4.8 & 3.06 & 0.92 & 0.85 & 0.10 \\
TiCodec \textcolor[gray]{0.6}{1:1} & 18.9 & 13.2 & 2.69 & 0.86 & 0.81 & 0.15 \\
Mimi \textcolor[gray]{0.6}{1:8} & 7.7 & 4.5 & 2.94 & 0.94 & 0.83 & 0.11 \\
Mimi \textcolor[gray]{0.6}{1:4} & 12.7 & 8.0 & 2.48 & 0.86 & 0.81 & 0.15 \\
Mimi \textcolor[gray]{0.6}{1:2} & 22.9 & 16.9 & 1.86 & 0.73 & 0.56 & 0.26 \\
DualCodec \textcolor[gray]{0.6}{1:8}$^\dagger$ & 5.0 & 2.8 & 3.67 & 0.99 & 0.81 & 0.09 \\
DualCodec \textcolor[gray]{0.6}{1:4}$^\dagger$ & 5.5 & 3.1 & 3.64 & 0.97 & 0.83 & 0.10 \\
DualCodec \textcolor[gray]{0.6}{1:2}$^\dagger$ & 7.8 & 4.4 & 3.24 & 0.96 & 0.83 & 0.11 \\
X-Codec 2$^\dagger$ & 5.4 & 2.9 & 3.59 & 0.98 & 0.89 & 0.10 \\
BiCodec & 5.7 & 3.1 & 3.73 & 0.98 & 0.86 & 0.10 \\
WavTokenizer & 18.2 & 11.3 & 2.92 & 0.88 & 0.82 & 0.14 \\
StableCodec & 25.0 & 16.5 & 3.83 & 0.91 & 0.90 & 0.10 \\
Kanade 12.5Hz & 12.2 & 7.2 & 3.77 & 0.94 & 0.70 & 0.21 \\
Kanade 25Hz & 5.6 & 3.0 & 3.72 & 0.97 & 0.84 & 0.17 \\
\cmidrule(lr){1-7}
\multicolumn{7}{c}{ERJ~\citep{nakagawa2007erj} \textit{(accented speech)}} \\
\noalign{\vskip -1pt}
\cmidrule(lr){1-7}
Ground Truth & 14.9 & 8.0 & 3.73 & \NA & \NA & \NA \\
FACodec \textcolor[gray]{0.6}{1:6} & 18.2 & 9.9 & 3.73 & 0.98 & 0.90 & 0.06 \\
FACodec \textcolor[gray]{0.6}{1:3} & 22.0 & 12.3 & 3.37 & 0.97 & 0.81 & 0.09 \\
PAST \textcolor[gray]{0.6}{1:8} & 25.3 & 14.1 & 3.65 & 0.97 & 0.85 & 0.07 \\
PAST \textcolor[gray]{0.6}{1:4} & 33.7 & 19.4 & 3.04 & 0.92 & 0.75 & 0.12 \\
PAST \textcolor[gray]{0.6}{1:2} & 47.3 & 27.8 & 2.00 & 0.79 & 0.30 & 0.30 \\
ST \textcolor[gray]{0.6}{1:8} & 19.6 & 10.8 & 3.48 & 0.97 & 0.89 & 0.06 \\
ST \textcolor[gray]{0.6}{1:4} & 28.5 & 15.7 & 3.11 & 0.94 & 0.82 & 0.09 \\
ST \textcolor[gray]{0.6}{1:2} & 47.5 & 27.1 & 1.97 & 0.84 & 0.66 & 0.13 \\
TiCodec \textcolor[gray]{0.6}{1:4} & 18.3 & 10.3 & 3.29 & 0.96 & 0.88 & 0.07 \\
TiCodec \textcolor[gray]{0.6}{1:2} & 26.7 & 15.8 & 3.13 & 0.94 & 0.82 & 0.08 \\
TiCodec \textcolor[gray]{0.6}{1:1} & 47.9 & 30.1 & 2.82 & 0.92 & 0.80 & 0.10 \\
Mimi \textcolor[gray]{0.6}{1:8} & 27.3 & 17.1 & 2.84 & 0.95 & 0.84 & 0.08 \\
Mimi \textcolor[gray]{0.6}{1:4} & 45.5 & 30.6 & 2.31 & 0.90 & 0.74 & 0.11 \\
Mimi \textcolor[gray]{0.6}{1:2} & 67.5 & 49.0 & 1.70 & 0.73 & 0.46 & 0.22 \\
DualCodec \textcolor[gray]{0.6}{1:8}$^\dagger$ & 17.1 & 9.4 & 3.71 & 0.98 & 0.86 & 0.07 \\
DualCodec \textcolor[gray]{0.6}{1:4}$^\dagger$ & 21.5 & 11.9 & 3.66 & 0.96 & 0.83 & 0.08 \\
DualCodec \textcolor[gray]{0.6}{1:2}$^\dagger$ & 29.2 & 16.7 & 3.25 & 0.94 & 0.80 & 0.09 \\
X-Codec 2$^\dagger$ & 20.7 & 11.3 & 3.69 & 0.97 & 0.86 & 0.08 \\
BiCodec & 21.4 & 11.7 & 3.76 & 0.97 & 0.86 & 0.07 \\
WavTokenizer & 51.7 & 31.6 & 3.06 & 0.91 & 0.82 & 0.08 \\
StableCodec & 51.4 & 29.3 & 4.03 & 0.91 & 0.87 & 0.06 \\
Kanade 12.5Hz & 33.8 & 18.6 & 3.78 & 0.95 & 0.80 & 0.09 \\
Kanade 25Hz & 22.9 & 12.3 & 3.75 & 0.96 & 0.86 & 0.07 \\
\bottomrule
\end{tabular}
\end{table}

\end{document}